\documentclass[runningheads]{llncs}
\usepackage{graphicx}
\usepackage{comment}
\usepackage{amsmath,amssymb} 
\usepackage{color}

\usepackage{soul}
\usepackage{capt-of}
\usepackage{multicol}
\usepackage{multirow}

\usepackage{array}
\newcommand{\PreserveBackslash}[1]{\let\temp=\\#1\let\\=\temp}
\newcolumntype{C}[1]{>{\PreserveBackslash\centering}p{#1}}
\newcolumntype{R}[1]{>{\PreserveBackslash\raggedleft}p{#1}}
\newcolumntype{L}[1]{>{\PreserveBackslash\raggedright}p{#1}}

\newcommand{\figref}[1]{Fig.~\ref{#1}}
\newcommand{\figsref}[1]{Figs.~\ref{#1}}
\newcommand{\tabref}[1]{Tab.~\ref{#1}}

\newcommand{\eqnref}[1]{Eq.~(\ref{#1})}

\newcommand{\secref}[1]{Sec.~\ref{#1}}

\newcommand{\Figref}[1]{Figure~\ref{#1}}

\newcommand{\Tabref}[1]{Table~\ref{#1}}

\newcommand{\ie}{\textit{i.e.}}
\newcommand{\eg}{\textit{e.g.}}
\newcommand{\etal}{\textit{et al.}}

\newcommand{\csmall}{\fontsize{8}{9.5}\selectfont}
\newcommand{\cfoot}{\fontsize{7}{8}\selectfont}
\newcommand{\ctiny}{\fontsize{5.5}{7}\selectfont}

\newcommand{\absSum}{$\mathtt{Abs{-}Sum}$}
\newcommand{\absSumStar}{$\mathtt{Abs{-}Sum^{*}}$}
\newcommand{\tanhC}{$\mathtt{Tanh{-}C}$}
\newcommand{\tgAbsSumStar}{$\mathtt{Tanh{-}\gamma{-}Abs{-}Sum^{*}}$}

\begin{document}
\pagestyle{headings}
\mainmatter
\def\ECCVSubNumber{1810}  

\title{Non-Local Spatial Propagation Network for Depth Completion} 


\titlerunning{Non-Local Spatial Propagation Network for Depth Completion}
%
\author{
Jinsun Park\inst{1} \and
Kyungdon Joo\inst{2} \and
Zhe Hu\inst{3} \and
Chi-Kuei Liu\inst{3} \and
In So Kweon\inst{1}
}
\authorrunning{J. Park et al.}
%
\institute{
Korea Advanced Institute of Science and Technology, Republic of Korea \\
\email{\{zzangjinsun, iskweon77\}@kaist.ac.kr} \and
Robotics Institute, Carnegie Mellon University \\
\email{kjoo@andrew.cmu.edu} \and
Hikvision Research America
}

\maketitle

\begin{abstract}
   In this paper, we propose a robust and efficient end-to-end non-local spatial propagation network for depth completion. 
   The proposed network takes RGB and sparse depth images as inputs and estimates non-local neighbors and their affinities of each pixel, as well as an initial depth map with pixel-wise confidences. 
   The initial depth prediction is then iteratively refined by its confidence and non-local spatial propagation procedure based on the predicted non-local neighbors and corresponding affinities. 
   Unlike previous algorithms that utilize fixed-local neighbors, the proposed algorithm effectively avoids irrelevant local neighbors and concentrates on relevant non-local neighbors during propagation. 
   In addition, we introduce a learnable affinity normalization to better learn the affinity combinations compared to conventional methods. 
   The proposed algorithm is inherently robust to the mixed-depth problem on depth boundaries, which is one of the major issues for existing depth estimation/completion algorithms. 
   Experimental results on indoor and outdoor datasets demonstrate that the proposed algorithm is superior to conventional algorithms in terms of depth completion accuracy and robustness to the mixed-depth problem.
   Our implementation is publicly available on the project page.\footnote{\texttt{https://github.com/zzangjinsun/NLSPN\_ECCV20}}
\end{abstract}

\section{Introduction}
\label{sec:intro}


\begin{figure}[t]
\begin{center}
\begin{tabular}{@{}c@{\hskip 0.001\linewidth}c@{\hskip 0.001\linewidth}c@{\hskip 0.001\linewidth}c@{\hskip 0.001\linewidth}c}
\includegraphics[width=0.199\linewidth]{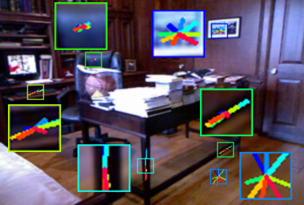} &
\includegraphics[width=0.199\linewidth]{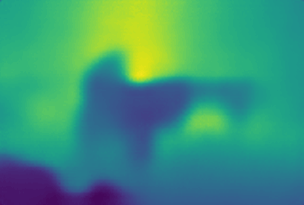} &
\includegraphics[width=0.199\linewidth]{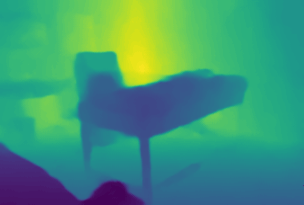} &
\includegraphics[width=0.199\linewidth]{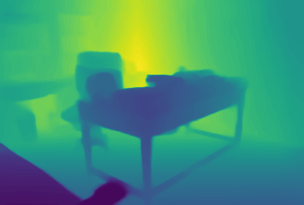} &
\includegraphics[width=0.199\linewidth]{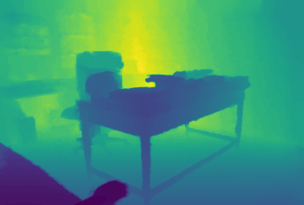} \\
{\csmall (a)} & {\csmall (b)} & {\csmall (c)} & {\csmall (d)} & {\csmall (e)}
\end{tabular}
\caption{
\textbf{Example of the depth completion on the NYU Depth V2 dataset~\cite{silberman2012indoor}}. 
(a)~RGB image and a few samples of the estimated non-local neighbors. Depth completion results by (b)~direct regression~\cite{ma2018sparse}, (c)~local propagation~\cite{cheng2018depth}, and (d)~non-local propagation (ours), respectively, and (e)~the ground truth. 
}
\label{fig:intro}
\end{center}
\end{figure}

%
Depth estimation has become an important problem in recent years with the rapid growth of computer vision applications, such as augmented reality, unmanned aerial vehicle control, autonomous driving, and motion planning. 
To obtain a reliable depth prediction, information from various sensors is utilized, \eg, RGB cameras, radar, LiDAR, and ultrasonic sensors~\cite{tesla_autopilot,uber_atg}. 
Depth sensors, such as LiDAR sensors, produce accurate depth measurements with high frequency. 
However, the density of the acquired depth is often sparse due to hardware limitations, such as the number of scanning channels. 
To overcome such limitations, there have been a lot of works to estimate dense depth information based on the given sparse depth values, called \textit{depth completion}.

Early methods for depth completion~\cite{uhrig2017sparsity,chodosh2018deep} rely only on sparse measurement. Therefore, their predictions suffer from unwanted artifacts, such as blurry and mixed-depth values (\ie, mixed-depth problem).
Because RGB images show subtle changes of color and texture, recent methods use RGB images as the guidance to predict accurate dense depth maps.

Direct depth completion algorithms~\cite{uhrig2017sparsity,ma2018sparse} take RGB or RGB-D images and directly infer a dense depth using a deep convolutional neural network~(CNN). 
These direct algorithms have shown superior performance compared to conventional ones; however, they still generate blurry depth maps near depth boundaries. 
Soon after, this phenomenon is alleviated by recent affinity-based spatial propagation methods~\cite{cheng2018depth,xu2019depth}. 
By learning affinities for local neighbors and iteratively refining depth predictions, the final dense depth becomes more accurate. 
%
Nonetheless, previous propagation networks~\cite{liu2017learning,cheng2018depth} have an explicit limitation that they have a fixed-local neighborhood configuration for propagation. 
Fixed-local neighbors often have irrelevant information that should not be mixed with reference information, especially on depth boundaries. 
Hence, they still suffer from the mixed-depth problem in the depth completion task (see \figref{fig:intro}(c)).

To tackle the problem, we propose a Non-Local Spatial Propagation Network (NLSPN) that predicts non-local neighbors for each pixel (\ie, where the information should come from) and then aggregates relevant information using the spatially-varying affinities (\ie, how much information should be propagated), which are also predicted from the network. 
%
%
By relaxing the fixed-local neighborhood configuration, the proposed network can avoid irrelevant local neighbors affiliated with other adjacent objects. 
%
Therefore, our method is inherently robust to the mixed-depth problem. 
In addition, based on our analysis of conventional affinity normalization schemes, we propose a learnable affinity normalization method that has a larger representation capability of affinity combinations. 
It enables more accurate affinity estimation and thus improves the propagation among non-local neighbors. 
To further improve robustness to outliers from input and inaccurate initial prediction, we predict the confidence of the initial dense depth simultaneously, and it is incorporated into the affinity normalization to minimize the propagation of unreliable depth values. 
Experimental results on the indoor~\cite{silberman2012indoor} and outdoor~\cite{uhrig2017sparsity} datasets demonstrate that our method achieves superior depth completion performance compared with state-of-the-art methods.

\section{Related Work}
\label{sec:related}

\noindent\textbf{Depth Estimation and Completion} \ 
The objective of depth estimation is to generate dense depth predictions based on various input information, such as a single RGB image, multi-view images, sparse LiDAR measurements, and so on. 
%
Conventional depth estimation algorithms often utilize information from a single modality. 
Eigen~\etal~\cite{eigen2014depth} used a multi-scale neural network to predict depth from a single image. 
In the method introduced by Zbontar and LeCun~\cite{zbontar2016stereo}, the deep features of image patches are extracted from stereo rectified images, and then the disparity is determined by searching for the most similar patch along the epipolar line. 
Depth estimation with accurate but sparse depth information (\ie, depth completion) has been intensively explored as well. 
Uhrig~\etal~\cite{uhrig2017sparsity} proposed sparsity invariant CNNs to predict a dense depth map given a sparse depth image from a LiDAR sensor. 
Ma and Sertac~\cite{ma2018sparse} introduced a method to construct a 4D volume by concatenating RGB and sparse depth images and then feed it into an encoder-decoder CNN for the final prediction. 
%
%
Chen~\etal~\cite{chen2019learning} adopted a fusion of 2D convolution and 3D continuous convolution to effectively consider the geometric configuration of 3D points.

\noindent\textbf{Spatial Propagation Network} \ 
Although direct depth completion algorithms have demonstrated decent performance, sparse-to-dense propagation with accurate guidance from different modalities (\eg, an RGB image) is a more effective way to obtain dense prediction from sparse inputs~\cite{cheng2018depth,xu2019depth,levin2006closed,park2017unified}. 
Liu~\etal~\cite{liu2017learning} proposed a spatial propagation network (SPN) to learn local affinities. 
The SPN learns task-specific affinity values from large-scale data, and it can be applied to a variety of high-level vision tasks, including depth completion and semantic segmentation. 
However, the individual three-way connection in four-direction is adopted for spatial propagation, which is not suitable for considering all local neighbors simultaneously. 
This limitation was overcome by Cheng~\etal~\cite{cheng2018depth}, who proposed a convolutional spatial propagation network (CSPN) to predict affinity values for local neighbors and update all the pixels simultaneously with their local context for efficiency. 
%
%
%
However, both the SPN and the CSPN rely on fixed-local neighbors, which could be from irrelevant objects. 
Therefore, the propagation based on those neighbors would result in mixed-depth values, and the iterative propagation procedure used in their architectures would increase the impact. 
Moreover, the fixed neighborhood patterns restrict the usage of relevant but wide-range (\textit{i.e.}, non-local) context within the image.

\noindent\textbf{Non-Local Network} \ 
The importance of non-local information has been widely explored in various vision tasks~\cite{buades2005non,wang2018non,yoon2006adaptive,shim2020robust}. 
%
%
Recently, a non-local block in deep neural networks was proposed by Wang~\etal~\cite{wang2018non}. 
It consists of pairwise affinity calculation and feature-processing modules. 
The authors demonstrated the effectiveness of non-local blocks by embedding them into existing deep networks for video classification and image recognition. 
These methods showed significant improvement over local methods.

\begin{figure*}[t]
\begin{center}
\includegraphics[width=0.998\linewidth]{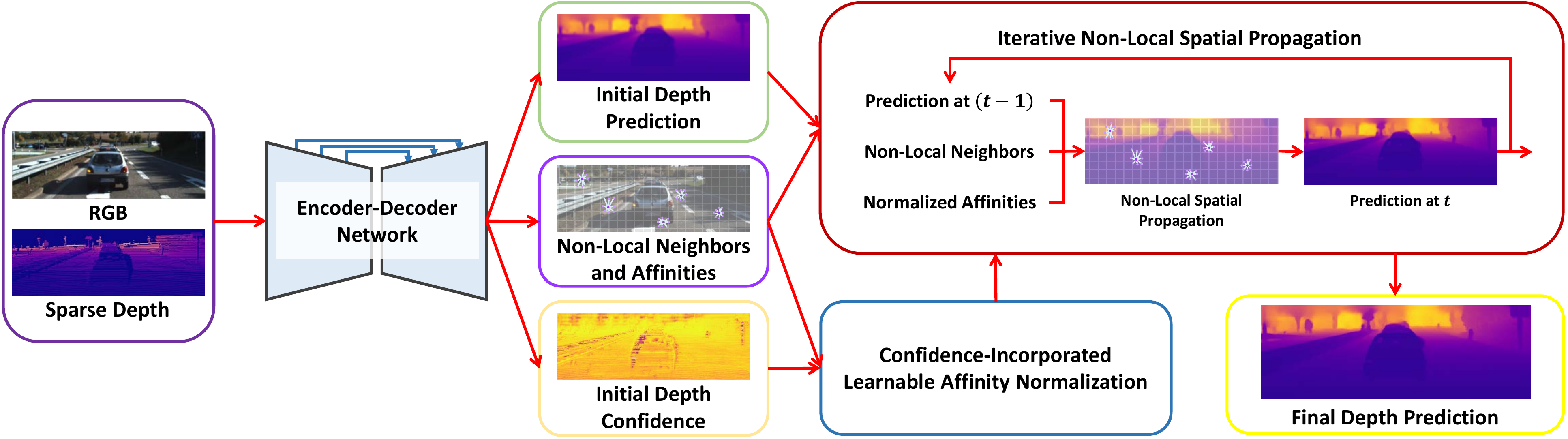}
\caption{
\textbf{Overview of the proposed algorithm}. 
The encoder-decoder network is built upon the residual network~\cite{he2016deep}. Given RGB and sparse depth images, an initial dense depth and its confidence, non-local neighbors, and corresponding affinities are predicted from the network. Then non-local spatial propagation is conducted iteratively with the confidence-incorporated learnable affinity normalization.
}
\label{fig:overview}
\end{center}
\end{figure*}

\noindent\textbf{Our Work} \ 
Unlike previous algorithms~\cite{liu2017learning,cheng2018depth,xu2019depth}, our network is trained to predict non-local neighbors with corresponding affinities. 
In addition, our learnable affinity normalization algorithm searches for the optimal affinity space, which has not been explored in conventional algorithms~\cite{chen2016semantic,liu2017learning,cheng2018depth}. 
Furthermore, we incorporate the confidence of the initial dense depth prediction (which will be refined by propagation procedure) into affinity normalization to minimize the propagation of unconfident depth values. 
%
\Figref{fig:overview} shows an overview of our algorithm. 
Each component will be described in subsequent sections in detail.


\section{Non-Local Spatial Propagation}
\label{sec:sp}

The goal of spatial propagation is to estimate missing values and refine less confident values by propagating neighbor observations with corresponding affinities (\textit{i.e.}, similarities). 
Spatial propagation has been utilized as one of the key modules in various computer vision applications~\cite{perona1990scale,levin2006closed,krahenbuhl2011efficient}. 
In particular, spatial propagation is suitable for the depth completion task~\cite{liu2017learning,cheng2018depth,xu2019depth}, and its superior performance compared to direct regression algorithms has been demonstrated~\cite{uhrig2017sparsity,ma2018sparse}. 
In this section, we first briefly review the local SPNs and their limitations, and then describe the proposed non-local SPN.

\subsection{Local Spatial Propagation Network}
\label{subsec:lspn}
Let $\mathbf{X}=(x_{m,n})\in\mathbb{R}^{{M{\times}N}}$ denote a 2D map to be updated by spatial propagation, where $x_{m,n}$ denotes the pixel value at $(m,n)$. 
The propagation of $x_{m,n}$ at the step $t$ with its local neighbors, denoted by $\mathcal{N}_{m,n}$, is defined as follows:
%
\begin{equation}
    x^{t}_{m,n} {=} w_{m,n}^{c}x^{t-1}_{m,n} {+} \sum_{(i,j) \in \mathcal{N}_{m,n}}\hspace{-2mm}w_{m,n}^{i,j}x^{t-1}_{i,j},
\label{eq:sp}
\end{equation}
where $(m,n)$ and $(i,j)$ are the coordinates of reference and neighbor pixels, respectively; 
$w_{m,n}^{c}$ represents the affinity of the reference pixel; and $w_{m,n}^{i,j}$ indicates the affinity between the  pixels at $(m,n)$ and $(i,j)$. 
%
The first term in the right-hand side represents the propagation of the reference pixel, while the second term stands for the propagation of its neighbors weighted by the corresponding affinities. 
The affinity of the reference pixel $w_{m,n}^{c}$ (\ie, how much the original value will be preserved) is obtained as 
%
\begin{equation}
     w_{m,n}^{c} = 1 - \sum_{(i,j) \in \mathcal{N}_{m,n}}w^{i,j}_{m,n}.
    \label{eq:aff_norm_ref}
\end{equation}

\begin{figure}[t]
    \begin{center}
    \begin{tabular}{@{}c@{\hskip 0.008\linewidth}c@{\hskip 0.008\linewidth}c@{\hskip 0.008\linewidth}c@{\hskip 0.008\linewidth}c@{\hskip 0.008\linewidth}c}
    \includegraphics[width=0.16\linewidth]{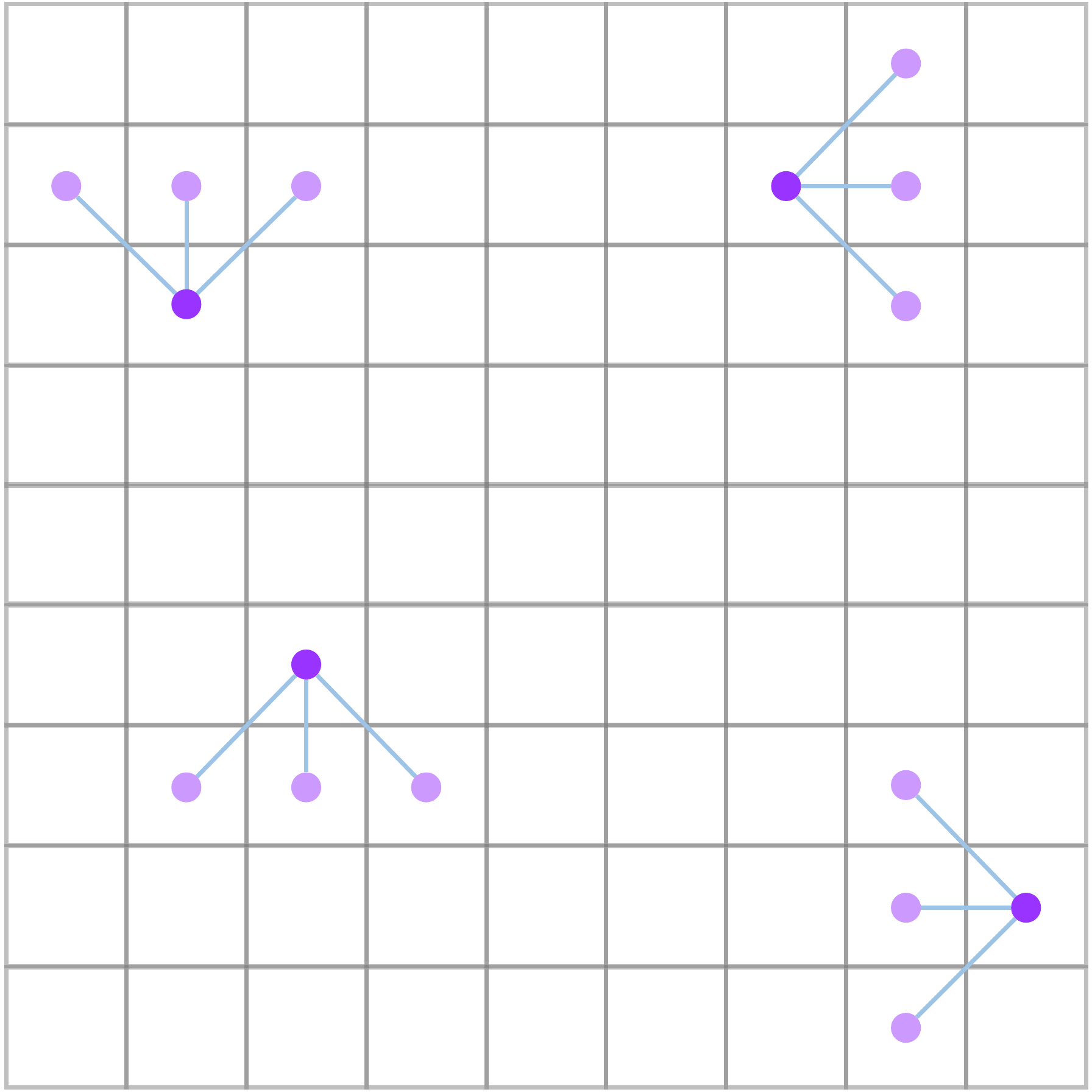} & 
    \includegraphics[width=0.16\linewidth]{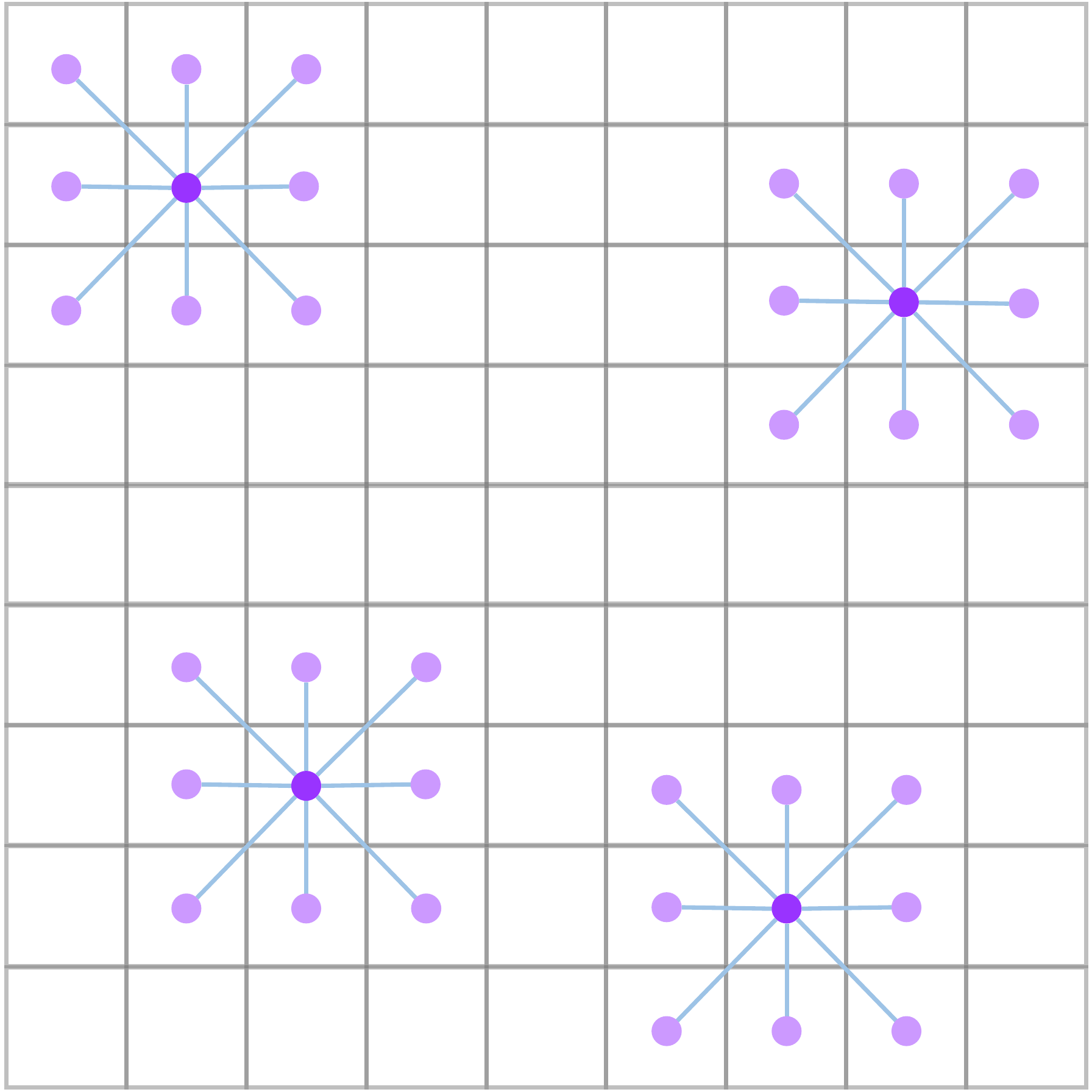} & 
    \includegraphics[width=0.16\linewidth]{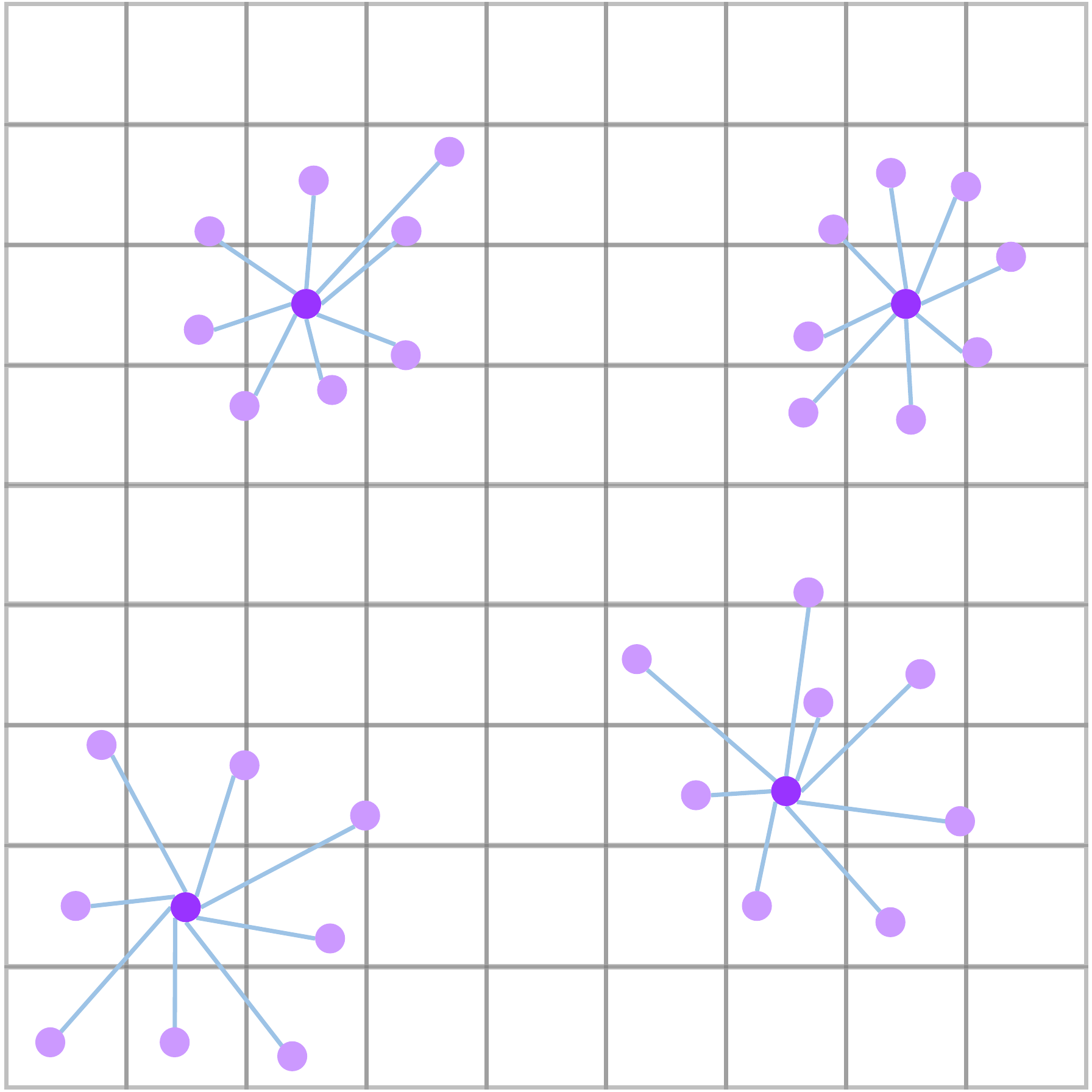} &
    \includegraphics[width=0.16\linewidth]{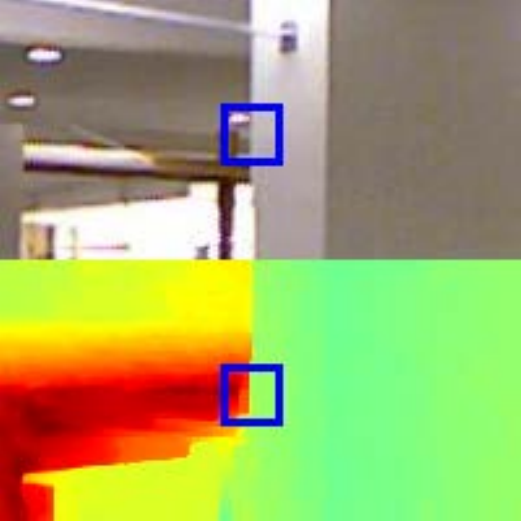} & 
    \includegraphics[width=0.16\linewidth]{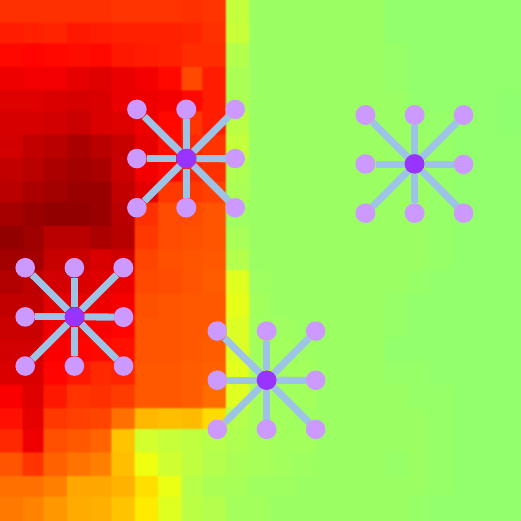} & 
    \includegraphics[width=0.16\linewidth]{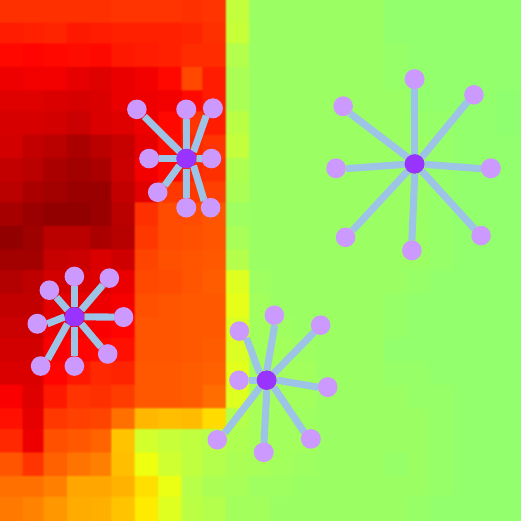} \\
    {\cfoot (a) SPN~\cite{liu2017learning}} & {\cfoot (b) CSPN~\cite{cheng2018depth}} & {\cfoot (c) Ours} &
    {\cfoot (d) RGB/Depth} & {\cfoot (e) Fixed-local} & {\cfoot (f) Non-local}
    \end{tabular}
    \caption{\textbf{Visual comparison of SPNs}.
    (a)-(c) Examples of neighbor configurations of the (a)~SPN~\cite{liu2017learning}, (b)~CSPN~\cite{cheng2018depth}, and (c)~NLSPN (ours), where purple and light purple pixels denote reference and neighboring pixels, respectively. 
    Compared to the others, our neighbor configuration is highly flexible, and can be fractional. 
    (d)-(f)~Comparison of fixed-local and non-local configurations for various situations. The fixed-local configuration~(e) cannot utilize relevant information beyond the fixed-local region. 
    In contrast, the non-local configuration~(f) avoids this problem effectively by predicting and utilizing relevant neighbors at various distances without limitation.
    }
    \label{fig:neighbors}
    \end{center}
\end{figure}

\noindent\textbf{Spatial Propagation Network} \ 
The original SPN~\cite{liu2017learning} is formulated on the configuration of three-way local connections, where each pixel is linked to three adjacent pixels from the previous row or column~(see \figref{fig:neighbors}(a)). 
%
%
For instance, the local neighbors of the pixel at $(m, n)$ for top-to-bottom propagation (\ie, vertical) in the SPN, denoted by $\mathcal{N}^{\mathrm{S}}_{m,n}$, are defined as follows: 
%
\begin{equation}
    \mathcal{N}^{\mathrm{S}}_{m,n} = \left\{x_{m+p,n+q}\ |\ p = -1, q \in \left\{-1, 0, 1\right\} \right\}.
\label{eq:spn_neighbor}
\end{equation}
The local neighbors for other directions (\textit{i.e.}, bottom-to-top, left-to-right and right-to-left) can be defined in similar ways. 
\Figref{fig:neighbors}(a) shows several examples of $\mathcal{N}^{\mathrm{S}}$ for other directions. 
Note that the SPN updates rows or columns in $\mathbf{X}$ sequentially. 
%
%
Thus, a natural limitation of the three-way connection is that it does not explore information from all the directions simultaneously.

\noindent\textbf{Convolutional Spatial Propagation Network} \ 
To consider all the possible propagation directions together, the original SPN propagates in four directions individually. 
Then it utilizes max-pooling to integrate those predictions~\cite{liu2017learning}. 
%
The CSPN~\cite{cheng2018depth} addresses the inefficiency issue by simplifying separate propagations via convolution operation at each propagation step. 
For the CSPN with a $3{\times}3$ local window size, the local neighbors $\mathcal{N}^{\mathrm{CS}}_{m,n}$ are defined as follows: 
%
\begin{equation}
    \mathcal{N}^{\mathrm{CS}}_{m,n} = \{x_{m+p,n+q}\ |\ p \in \left\{-1, 0, 1\right\}, q \in \left\{-1, 0, 1\right\}, (p,q) \neq (0, 0) \}.
\label{eq:cspn_neighbor}
\end{equation}
\Figref{fig:neighbors}(b) shows some examples of $\mathcal{N}^{\mathrm{CS}}$. 
For more details of each network (the SPN and the CSPN), please refer to earlier works~\cite{liu2017learning,cheng2018depth}.

\subsection{Non-Local Spatial Propagation Network}
\label{subsec:nlspn}

The SPN and the CSPN are effective in propagating information from more confident areas into less confident ones with data-dependent affinities. 
However, their potential improvement is inherently limited by the fixed-local neighborhood configuration (\figref{fig:neighbors}(e)).
%
%
The fixed-local neighborhood configuration ignores object/depth distribution within the local area; thus, it often results in mixed-depth values of foreground and background objects after propagation. 
Although affinities predicted from the network can alleviate the depth mixing between irrelevant pixels to a certain degree, they can hardly avoid incorrect predictions and hold up the use of appropriate neighbors beyond the local area.

To resolve the above issues, we introduce a deep neural network that estimates the neighbors of each pixel beyond the local region (\ie, non-local) based on color and depth information within a wide area. 
%
%
The non-local neighbors $\mathcal{N}^{\mathrm{NL}}_{m,n}$ are defined as follows: 
%
\begin{equation}
    \mathcal{N}^{\mathrm{NL}}_{m,n} = \{x_{m+p,n+q}\ |\ (p, q) \in f_{\phi}(\mathbf{I}, \mathbf{D}, m, n), \ p,q \in \mathbb{R} \},
\label{eq:nlspn_neighbor}
\end{equation}
%
where $\mathbf{I}$ and $\mathbf{D}$ are the RGB and sparse depth images, respectively, and $f_{\phi}(\cdot)$ is the non-local neighbor prediction network that estimates $K$ neighbors for each pixel, under the learnable parameters $\phi$.
We adopt an encoder-decoder CNN architecture for $f_{\phi}(\cdot)$, which will be described in \secref{subsec:net_arch}. 
It should be noted that $p$ and $q$ are real numbers in \eqnref{eq:nlspn_neighbor}; thus, the non-local neighbors can be defined to sub-pixel accuracy, as illustrated in \figref{fig:neighbors}(c).
%

\Figref{fig:neighbors}(f) shows some examples of appropriate and desired non-local neighbors near depth boundaries. 
%
%
In the fixed-local setup, affinity learning learns how to encourage the influence of the related pixels and suppress that of unrelated ones simultaneously.
On the contrary, affinity learning with the non-local setup concentrates on relevant neighbors, and this facilitates the learning process.

\section{Confidence-Incorporated Affinity Learning}
\label{sec:aff}

Affinity learning is one of the key components in SPNs, which enables accurate and stable propagation. 
Conventional affinity-based algorithms utilize color statistics or hand-crafted features~\cite{levin2006closed,saxena2006learning,krahenbuhl2011efficient}. 
Recent affinity learning methods~\cite{liu2015deep,liu2017learning,cheng2018depth} adopt deep neural networks to predict affinities and show substantial performance improvement.
In these methods, affinity normalization plays an important role to stabilize the propagation process. 

In this section, we analyze the conventional normalization approach and its limitation, and then propose a normalization approach in a learnable way. 
Moreover, we incorporate the confidence of the initial prediction during normalization to suppress negative effects from unreliable depth values during propagation.

\subsection{Affinity Normalization}
\label{subsec:aff_norm}

The purpose of affinity normalization is to ensure stability during propagation. 
For stability, the norm of the temporal Jacobian of $x$, $\partial x^{t} / \partial x^{t-1}$ should be equal to or less than one~\cite{liu2017learning}. 
Under the spatial propagation formulation in \eqnref{eq:sp}, this condition would be satisfied if $\sum_{(i,j) \in \mathcal{N}_{m,n}}|w^{i,j}_{m,n}| \leq 1, \ \forall m, n$. 
To enforce the condition, previous works~\cite{liu2017learning,cheng2018depth} normalize affinities by the absolute-sum (dubbed \absSum) as follows: 
%
\begin{equation}
    w^{i,j}_{m,n} = \hat{w}^{i,j}_{m,n} / \sum_{(i,j) \in \mathcal{N}_{m,n}}|\hat{w}^{i,j}_{m,n}|,
    \label{eq:aff_norm_orig}
\end{equation}
where $\hat{w}$ denotes the raw affinity before normalization. 
%
%
%
Although the stability condition is satisfied by \absSum, it has a problem in that the viable combinations of normalized affinities are biased to a narrow high-dimensional space.

Without loss of generality, we first analyze the biased affinity problem using a toy example of the 2-neighbor case and then present solutions to the issue. 
In the 2-neighbor case, we denote affinities of the two neighbors as $w_{1}$ and $w_{2}$ with a slight abuse of notation. 
We assume that the unnormalized affinities are sampled from the standard normal distribution, $\mathsf{N}(0, 1)$ for simplicity.

\begin{figure*}[t]
\begin{center}
\begin{tabular}{@{}c@{\hskip 0.005\linewidth}c@{\hskip 0.005\linewidth}c@{\hskip 0.005\linewidth}c@{\hskip 0.005\linewidth}c@{\hskip 0.005\linewidth}c}
\includegraphics[width=0.02\linewidth]{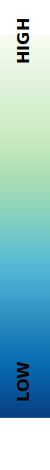} & 
\includegraphics[width=0.19\linewidth]{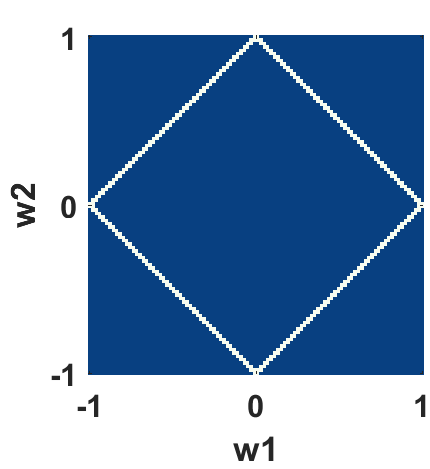} & 
\includegraphics[width=0.19\linewidth]{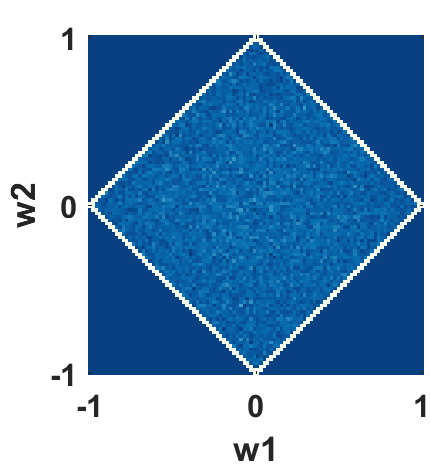} & 
\includegraphics[width=0.19\linewidth]{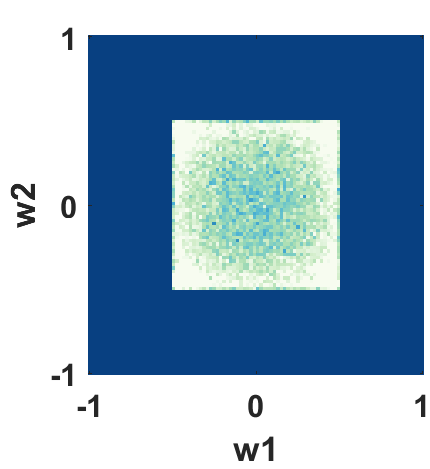} & 
\includegraphics[width=0.19\linewidth]{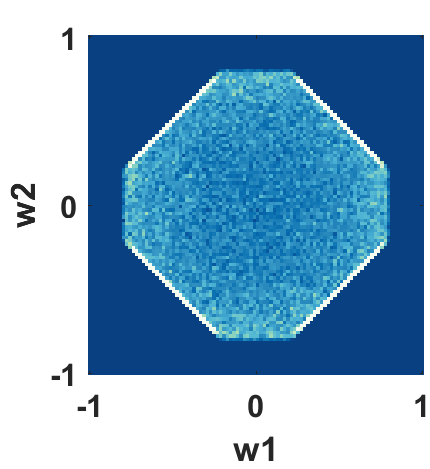} & 
\includegraphics[width=0.19\linewidth]{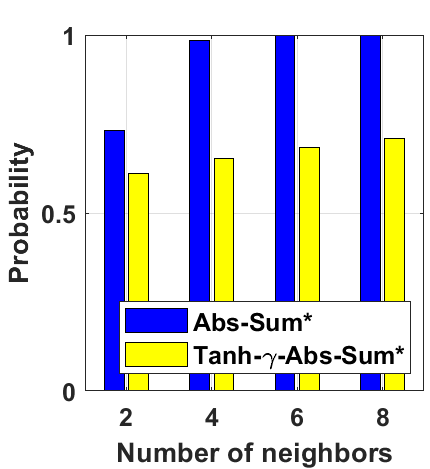} \\
 & {\cfoot (a) \absSum} & {\cfoot (b) \absSumStar} & {\cfoot (c) \tanhC} &
{\cfoot (d) \resizebox*{0.16\hsize}{!}{\tgAbsSumStar}} & {\cfoot (e) Norm. Prob.}
\end{tabular}
\caption{\textbf{Illustration of affinity normalization schemes}.
(a)-(d)~Affinity distribution after various normalization schemes for the 2-neighbor case. Color bar is shown on the left. 
(e)~Probabilities of normalization with different strategies for each number of neighbors. Please refer to the text for details.
}
\label{fig:aff_norm}
\end{center}
\end{figure*}

For the \absSum, the normalized affinities lie on the lines satisfying $|w_{1}| + |w_{2}| = 1$ (referred to as $A_{1}$), as shown in \figref{fig:aff_norm}(a).
%
This limits the usage of potentially advantageous affinity configuration within the area $|w_{1}| + |w_{2}| < 1$ (referred to as $A_{2}$). 
%
%
To fully explore the affinity configuration $|w_{1}| + |w_{2}| \leq 1$, a simple remedy is to apply \eqnref{eq:aff_norm_orig} only when $\sum_{i}|w_{i}| > 1$ (noted as \absSumStar). 
\Figref{fig:aff_norm}(b) shows the affinity distribution of our simple remedy. 
However, the affinities normalized by \absSumStar~still have a high chance to fall on $A_{1}$. 
Indeed, with the increasing number of neighbors $K$, the affinities are more likely to lie on $A_{1}$. (\eg, the normalization probability is $0.985$ when $K=4$).
%
%
\Figref{fig:aff_norm}(e) (blue bars) shows the probability of affinities falling on $A_1$ with various $K$ values.
%

One way to reduce the bias is to limit the range of raw affinities~\cite{liu2016learning}, for example, to $[-1/C, 1/C]$ using the hyperbolic tangent function $(\texttt{tanh}(\cdot))$ with a normalization factor $C$. 
We refer to this normalization procedure as \tanhC, which is defined as follows: 
%
\begin{equation}
    w^{i,j}_{m,n} = \texttt{tanh}(\hat{w}^{i,j}_{m,n}) / C, \qquad C \geq K, 
\label{eq:tanh_const}
\end{equation}
where the condition $C \geq K$ enforces the normalized affinities to guarantee $\sum_{(i,j) \in \mathcal{N}_{m,n}}|w^{i,j}_{m,n}| \leq 1$; therefore, this condition ensures stability. 
%
%
\Figref{fig:aff_norm}(c) shows the affinity distribution of \tanhC~when $C = 2$. 
With a sacrifice of boundary values, \tanhC~enables a more balanced affinity distribution. 
%
%
%
Moreover, the optimal value of $C$ in \tanhC~may vary depending on the training task, \eg, the number of neighbors, the activation functions, and the dataset.

To determine the optimal value for the task, we propose to learn the normalization factor together with non-local affinities, and apply the normalization only when $\sum_{(i,j) \in \mathcal{N}_{m,n}}|w^{i,j}_{m,n}| > 1$. 
The affinity of the proposed normalization, referred to as \tgAbsSumStar, is defined as follows: 
%
\begin{equation}
    w^{i,j}_{m,n} = \texttt{tanh}(\hat{w}^{i,j}_{m,n}) / \gamma,\ \ \gamma_{min} \leq \gamma \leq \gamma_{max},
\label{eq:tanh_gamma}
\end{equation}
where $\gamma$ denotes the learnable normalization parameter, and $\gamma_{min}$ and $\gamma_{max}$ are the minimum and maximum values that can be empirically set.
%
%
\Figref{fig:aff_norm}(d) shows an example of \tgAbsSumStar~when $\gamma = 1.25$. 
Here, \tgAbsSumStar~can be viewed as a mixture of \absSumStar~and \tanhC~(see \figsref{fig:aff_norm}(b) and (c)). 
The probability of affinities falling on the boundary with respect to the number of neighbors with $\gamma = K/2$ is shown in \figref{fig:aff_norm}(e) (yellow bars). 
Compared to \absSumStar, \tgAbsSumStar~still has a chance to avoid normalization, and it allows us to explore more diverse affinities with a larger number of neighbors.

\subsection{Confidence-Incorporated Affinity Normalization}
\label{subsec:conf_aff_norm}

\begin{figure*}[t]
\begin{center}
\begin{tabular}{@{}c@{\hskip 0.01\linewidth}c@{\hskip 0.01\linewidth}c@{\hskip 0.01\linewidth}c@{\hskip 0.01\linewidth}c}
\includegraphics[width=0.192\linewidth]{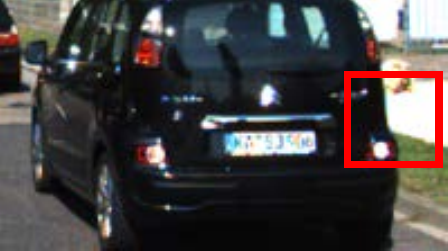} &
\includegraphics[width=0.192\linewidth]{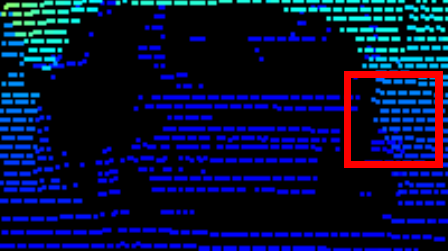} &
\includegraphics[width=0.192\linewidth]{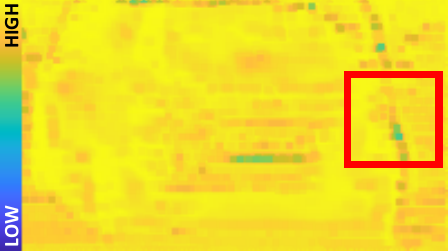} &
\includegraphics[width=0.192\linewidth]{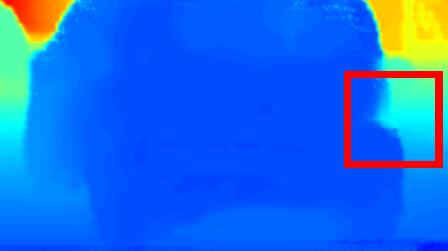} &
\includegraphics[width=0.192\linewidth]{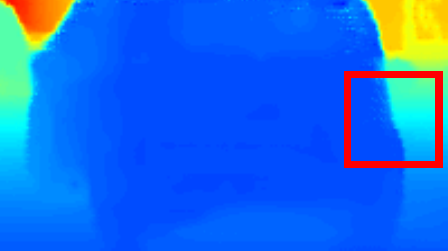} \\
{\csmall (a) RGB} & {\csmall (b) Depth} & {\csmall (c) Confidence} & {\csmall (d) Without Conf.} & {\csmall (e) With Conf.}
\end{tabular}
\caption{\textbf{Example of propagation with and without confidence incorporation}.
}
\label{fig:aff_conf_norm}
\end{center}
\end{figure*}

In the existing propagation frameworks~\cite{levin2006closed,saxena2006learning,krahenbuhl2011efficient,liu2015deep,liu2017learning,cheng2018depth}, the affinity depicts the correlation between pixels and provides guidance for propagation based on similarity. 
In this case, each pixel in the map is treated equally without consideration of its reliability.
However, in the depth completion task, different pixels should be weighted based on their reliability. 
For example, information from unreliable pixels (\eg, noisy pixels and pixels on depth boundaries) should not be propagated into neighbors regardless of their affinity to the neighboring pixels.
The recent work DepthNormal~\cite{xu2019depth} addresses this problem with confidence prediction. 
%
It utilizes confidence as a mask for the weighted summation of input and prediction for seed point preservation. 
However, it does not fully prevent the propagation of incorrect depth values because weighted summation is conducted before each propagation separately.
%
%

In this work, we consider the confidence map of pixels and combine it with affinity normalization.
That is, we predict not only the initial dense depth but also its confidence, and then the confidence is incorporated into affinity normalization to reduce disturbances from unreliable depths during propagation. 
The affinity of the confidence-incorporated \tgAbsSumStar~is defined as follows:
%
\begin{equation}
     w^{i,j}_{m,n} = c^{i,j} \cdot \texttt{tanh}(\hat{w}^{i,j}_{m,n}) / \gamma,
    \label{eq:conf_aff_norm}
\end{equation}
where $c^{i,j} \in [0, 1]$ denotes the confidence of the pixel at $(i,j)$. 
%
%

\Figref{fig:aff_conf_norm}(d) shows an example of a confidence-agnostic depth estimation result. 
Some noisy input depth points generate unreliable depth values with low confidences (see \figref{fig:aff_conf_norm}(c)). 
Without using confidence, the noisy and less confident pixels would harm their neighbor pixels during propagation and lead to unpleasing artifacts (see \figref{fig:aff_conf_norm}(d)).
After the incorporation of confidence into normalization, our algorithm can successfully eliminate the impact of unconfident pixels and generate more accurate depth estimation, as shown in \figref{fig:aff_conf_norm}(e).


\section{Depth Completion Network}
\label{sec:implementation}
In this section, we describe network architecture and loss functions for network training. 
The proposed NLSPN mainly consists of two parts: (1)~an encoder-decoder architecture for the initial depth map, a confidence map and non-local neighbors prediction with their raw affinities, and (2)~a non-local spatial propagation layer with a learnable affinity normalization.

\subsection{Network Architecture}
\label{subsec:net_arch}

%
The encoder-decoder part of the proposed network is built upon residual networks~\cite{he2016deep}, and it extracts high-level features from RGB and sparse depth images. 
Additionally, we adopt the encoder-decoder feature connection strategy~\cite{ronneberger2015u,cheng2018depth} to simultaneously utilize low-level and high-level features.

In \figref{fig:overview}, we provide an overview of our algorithm. 
Features from the encoder-decoder network are shared for the initial dense depth, confidence, non-local neighbor, and raw affinity estimation. 
Then non-local spatial propagation is conducted in an iterative manner. 
As described in~\secref{subsec:nlspn}, non-local neighbors can have fractional coordinates. 
To better incorporate fractional coordinates into training, differentiable sampling~\cite{jaderberg2015spatial,zhu2019deformable} is adopted during propagation. 
We note that our non-local propagation can be efficiently calculated by deformable convolutions~\cite{zhu2019deformable}. 
Therefore, each propagation requires a simple forward step of deformable convolution with our affinity normalization. 
Please refer to the supplementary material for the detailed network configuration.

\subsection{Loss Function}
For accurate prediction of the dense depth map, we train our network with $\ell_{1}$ or $\ell_{2}$ loss as a reconstruction loss with the ground truth depth as follows: 
\begin{equation}
    L_{recon}(\mathbf{D}^{gt}, \mathbf{D}^{pred}) = \frac{1}{\left\vert \mathcal{V} \right\vert} \sum_{v \in \mathcal{V}}\left\vert~d_{v}^{gt} - d_{v}^{pred}~\right\vert^{\rho},
\label{eq:loss_recon}
\end{equation}
where $\mathbf{D}^{gt}$ is the ground truth depth; $\mathbf{D}^{pred}$ is the prediction from our algorithm; and $d_{v}$, $\mathcal{V}$, and $\left\vert \mathcal{V} \right\vert$ denote the depth values at pixel index $v$, valid pixels of $\mathbf{D}^{gt}$, and the number of valid pixels, respectively. 
Here, $\rho$ is set to 1 for $\ell_{1}$ loss and 2 for $\ell_{2}$ loss.
Note that we do not have any supervision on the confidence because there is no ground truth; therefore, it is indirectly trained based on $L_{recon}$.

\section{Experimental Results}
\label{sec:exp}

In this section, we first describe implementation details and the training environment. 
After that, quantitative and qualitative comparisons to previous algorithms on indoor and outdoor datasets are presented. 
%
We also present ablation studies to verify the effectiveness of each component of the proposed algorithm.

The proposed method was implemented using PyTorch~\cite{paszke2017automatic} with NVIDIA Apex~\cite{nvidia_apex} and trained with a machine equipped with Intel Xeon E5-2620 and 4 NVIDIA GTX 1080 Ti GPUs. 
For all our experiments, we adopted an ADAM optimizer with $\beta_{1} = 0.9$, $\beta_{2} = 0.999$, and the initial learning rate of 0.001. 
The network training took about 1 and 3 days on the NYU Depth V2~\cite{silberman2012indoor} and KITTI Depth Completion~\cite{uhrig2017sparsity} datasets, respectively. 
We adopted the ResNet34~\cite{he2016deep} as our encoder-decoder baseline network.
The number of non-local neighbors was set to 8 for a fair comparison to other algorithms using $3{\times}3$ local neighbors. 
The number of propagation steps was set to 18 empirically. 
%
Other training details will be described for each dataset individually.
For the quantitative evaluation, we utilized the following commonly used metrics~\cite{silberman2012indoor,ma2018sparse,cheng2018depth}:
%
\begin{multicols}{2}
\begin{itemize}
\setlength\itemsep{0mm}
\cfoot{
    \item RMSE (mm) : {\ctiny $\sqrt{\frac{1}{\left\vert \mathcal{V} \right\vert}\sum_{v \in \mathcal{V}}\left\vert\ d_{v}^{gt} - d_{v}^{pred}\ \right\vert^{2}}$}
    \item MAE (mm) : {\ctiny $\frac{1}{\left\vert \mathcal{V} \right\vert}\sum_{v \in \mathcal{V}}\left\vert\ d_{v}^{gt} - d_{v}^{pred}\ \right\vert$}
    \item iRMSE (1/km) : {\ctiny $\sqrt{\frac{1}{\left\vert \mathcal{V} \right\vert}\sum_{v \in \mathcal{V}}\left\vert\ 1/d_{v}^{gt} - 1/d_{v}^{pred}\ \right\vert^{2}}$}
    \item iMAE (1/km) : {\ctiny $\frac{1}{\left\vert \mathcal{V} \right\vert}\sum_{v \in \mathcal{V}}\left\vert\ 1/d_{v}^{gt} - 1/d_{v}^{pred}\ \right\vert$}
    \item REL : {\ctiny $\frac{1}{\left\vert \mathcal{V} \right\vert}\sum_{v \in \mathcal{V}}\left\vert\ (d_{v}^{gt} - d_{v}^{pred}) / d_{v}^{gt}\ \right\vert$}
    \item $\delta_{\tau}$ : Percentage of pixels satisfying \\
    \hspace*{0.1\linewidth} {\ctiny $\texttt{max}\left(\frac{d_{v}^{gt}}{d_{v}^{pred}}, \frac{d_{v}^{pred}}{d_{v}^{gt}}\right) < \tau$}
}
\end{itemize}
\end{multicols}

\subsection{NYU Depth V2}
\label{subsec:exp_nyu}

\begin{figure*}[t]
\begin{center}
\renewcommand{\arraystretch}{0.2}
\begin{tabular}{@{}c@{\hskip 0.001\linewidth}c@{\hskip 0.001\linewidth}c@{\hskip 0.001\linewidth}c@{\hskip 0.001\linewidth}c@{\hskip 0.001\linewidth}c}
\includegraphics[width=0.16\linewidth]{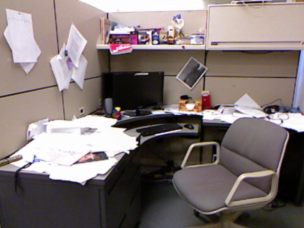} &
\includegraphics[width=0.16\linewidth]{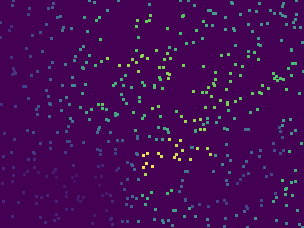} &
\includegraphics[width=0.16\linewidth]{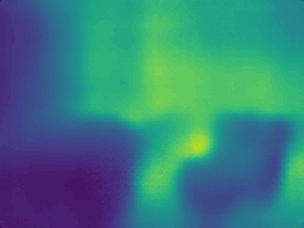} &
\includegraphics[width=0.16\linewidth]{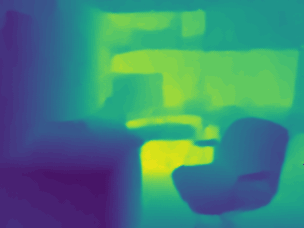} &
\includegraphics[width=0.16\linewidth]{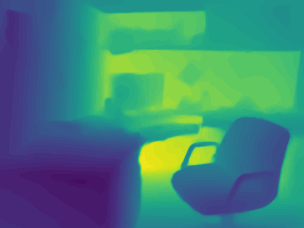} &
\includegraphics[width=0.16\linewidth]{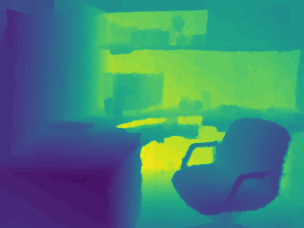} \\
\includegraphics[width=0.16\linewidth]{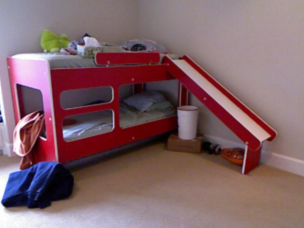} &
\includegraphics[width=0.16\linewidth]{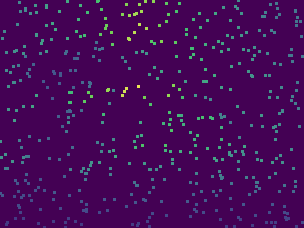} &
\includegraphics[width=0.16\linewidth]{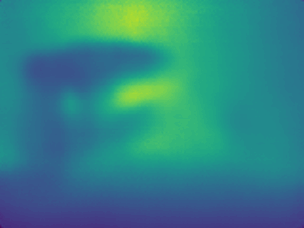} &
\includegraphics[width=0.16\linewidth]{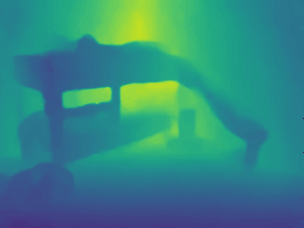} &
\includegraphics[width=0.16\linewidth]{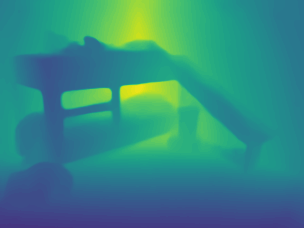} &
\includegraphics[width=0.16\linewidth]{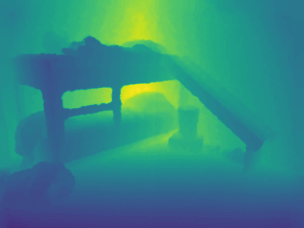} \\
{\cfoot (a) RGB} & {\cfoot (b) Depth} & {\cfoot (c) S2D~\cite{ma2018sparse}} &
{\cfoot (d) CSPN~\cite{cheng2018depth}} & {\cfoot (e) Ours} & {\cfoot (f) GT}
\end{tabular}
\caption{\textbf{Depth completion results on the NYUv2 dataset~\cite{silberman2012indoor}}. Note that sparse depth images are dilated for visualization.}
\label{fig:result_nyu}
\end{center}
\end{figure*}

The NYU Depth V2 dataset~\cite{silberman2012indoor} (NYUv2) consists of RGB and depth images of 464 indoor scenes captured by a Kinect sensor. 
For the training data, we utilized a subset of $\sim$50K images from the official training split. 
Each image was downsized to $320{\times}240$, and then $304{\times}228$ center-cropping was applied. 
We trained the model for 25 epochs with $\ell_{1}$ loss, and the learning rate decayed by 0.2 every 5 epochs after the first 10 epochs. 
We set the batch size to 24.
The official test split of 654 images was used for evaluation and comparisons.

In \figref{fig:result_nyu}, we present some depth completion results obtained for the NYUv2 dataset. 
As in previous works~\cite{ma2018sparse,cheng2018depth}, 500 depth pixels were randomly sampled from a dense depth image and used as the input along with the corresponding RGB image. 
For comparison, we provide results from the Sparse-to-Dense (S2D)~\cite{ma2018sparse} and the CSPN~\cite{cheng2018depth}. 
The S2D (\figref{fig:result_nyu}(c)) generates blurry depth images, as it is a direct regression algorithm. 
Compared to the S2D, the CSPN and our method generate depth maps with substantially improved accuracy thanks to the iterative spatial propagation procedure. 
However, the CSPN suffers from mixed-depth problems, especially on tiny or thin structures. 
In contrast, our method well preserves tiny structures and depth boundaries using non-local propagation.

\Tabref{tab:quan_nyu} shows the quantitative evaluation of the NYUv2 dataset. 
The proposed algorithm achieves the best result and outperforms other methods by a large margin (RMSE $0.020$m).
Compared to geometry-agnostic methods~\cite{ma2018sparse,liu2017learning,cheng2018depth}, geometry-aware ones~\cite{imran2019depth,cheng2019cspnpp,qiu2019deeplidar,xu2019depth} show better performance in general. 
%
%
The proposed algorithm can be also viewed as a geometry-aware algorithm because it implicitly explores geometrically relevant neighbors for propagation.

\begin{table}[t]
{\ctiny
    \begin{minipage}[t]{0.52\linewidth}
        \begin{center}
        \renewcommand{\arraystretch}{1.4}
        \begin{tabular}{C{19mm}|C{7mm}C{7mm}C{7mm}C{7mm}C{7mm}}
        \hline
        Method & RMSE (m) & REL & $\delta_{1.25}$ & $\delta_{1.25^{2}}$ & $\delta_{1.25^{3}}$ \\ \hline 
        S2D~\cite{ma2018sparse} & 0.230 & 0.044 & 97.1 & 99.4 & 99.8 \\ 
        \cite{ma2018sparse}+Bilateral~\cite{barron2016fast} & 0.479 & 0.084 & 92.4 & 97.6 & 98.9 \\ 
        \cite{ma2018sparse}+SPN~\cite{liu2017learning} & 0.172 & 0.031 & 98.3 & 99.7 & 99.9 \\ 
        DepthCoeff~\cite{imran2019depth} & 0.118 & 0.013 & 99.4 & \textbf{99.9} & - \\ 
        CSPN~\cite{cheng2018depth} & 0.117 & 0.016 & 99.2 & \textbf{99.9} & \textbf{100.0} \\ 
        CSPN++~\cite{cheng2019cspnpp} & 0.116 & - & - & - & - \\ 
        DeepLiDAR~\cite{qiu2019deeplidar} & 0.115 & 0.022 & 99.3 & \textbf{99.9} & \textbf{100.0} \\ 
        DepthNormal~\cite{xu2019depth} & 0.112 & 0.018 & 99.5 & \textbf{99.9} & \textbf{100.0} \\ 
        Ours & \textbf{0.092} & \textbf{0.012} & \textbf{99.6} & \textbf{99.9} & \textbf{100.0} \\ \hline
        \end{tabular}
        \caption{\textbf{Quantitative evaluation on the NYUv2~\cite{silberman2012indoor} dataset}. 
        Results are borrowed from each paper.
        Note that S2D~\cite{ma2018sparse} uses 200 sampled depth points per image as the input, while the others use 500. 
        }
        \label{tab:quan_nyu}
        \end{center}
    \end{minipage}
    \hfill
    \begin{minipage}[t]{0.45\linewidth}
        \begin{center}
        \renewcommand{\arraystretch}{1.4}
        \begin{tabular}{C{19mm}|C{8mm}C{8mm}C{7mm}C{7mm}}
        \hline
        Method & RMSE (mm) & MAE & iRMSE & iMAE \\ \hline 
        CSPN~\cite{cheng2018depth} & 1019.64 & 279.46 & 2.93 & 1.15 \\ 
        DDP~\cite{yang2019dense} & 832.94 & 203.96 & 2.10 & 0.85 \\ 
        NConv~\cite{eldesokey2019confidence} & 829.98 & 233.26 & 2.60 & 1.03 \\ 
        S2D~\cite{ma2018sparse} & 814.73 & 249.95 & 2.80 & 1.21 \\ 
        DepthNormal~\cite{xu2019depth} & 777.05 & 235.17 & 2.42 & 1.13 \\ 
        DeepLiDAR~\cite{qiu2019deeplidar} & 758.38 & 226.50 & 2.56 & 1.15 \\ 
        FuseNet~\cite{chen2019learning} & 752.88 & 221.19 & 2.34 & 1.14 \\ 
        CSPN++~\cite{cheng2019cspnpp} & 743.69 & 209.28 & 2.07 & 0.90 \\ 
        Ours & \textbf{741.68} & \textbf{199.59} & \textbf{1.99} & \textbf{0.84} \\ \hline
        \end{tabular}
        \caption{\textbf{Quantitative evaluation on the KITTI DC test dataset~\cite{uhrig2017sparsity}}.
        The results from other methods are obtained from the KITTI online evaluation site.}
        \label{tab:quan_kitti}
        \end{center}
    \end{minipage}
}
\end{table}

\subsection{KITTI Depth Completion}
\label{subsec:kitti_dc}
The KITTI Depth Completion (KITTI DC) dataset~\cite{uhrig2017sparsity} consists of over 90K RGB and LiDAR pairs. 
We ignored regions without LiDAR projection (\ie, top 100 pixels) and center-cropped $1216 \times 240$ patches for training. 
The proposed network was trained for 25 epochs with both $\ell_{1}$ and $\ell_{2}$ losses to balance RMSE and MAE, and the initial learning rate decayed by 0.4 every 5 epochs after the first 10 epochs. 
We used a batch size of 25 for the training.

\begin{figure*}[t]
\begin{center}
\renewcommand{\arraystretch}{0.2}
\begin{tabular}{@{}c@{\hskip 0.001\linewidth}c@{\hskip 0.001\linewidth}c@{\hskip 0.001\linewidth}c}
\raisebox{2\height}{{\csmall (a)}}  &
\includegraphics[width=0.32\linewidth]{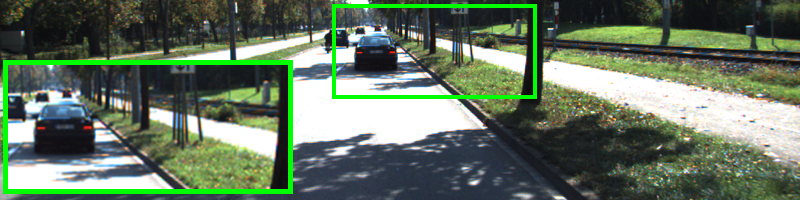} &
\includegraphics[width=0.32\linewidth]{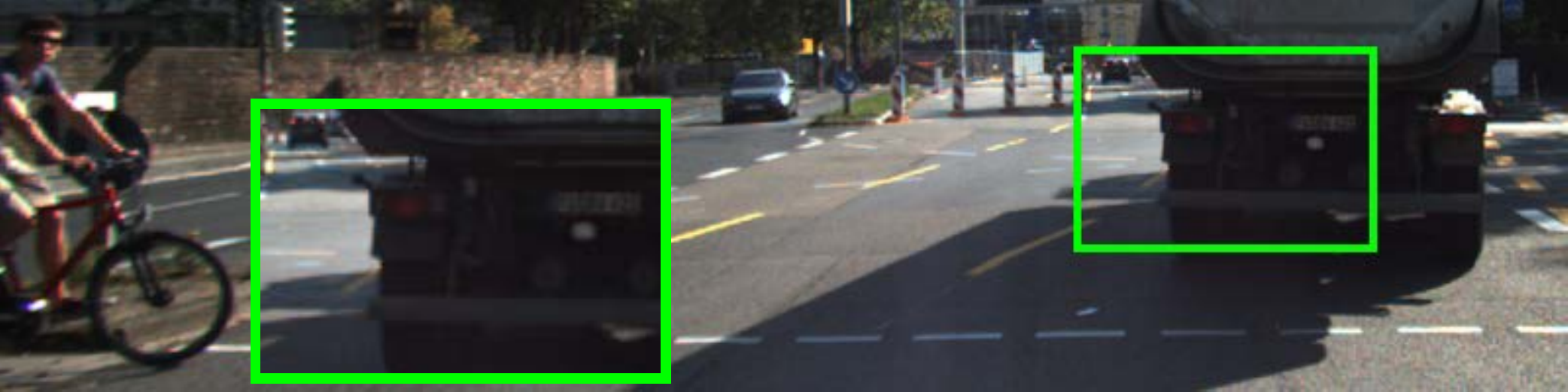} &
\includegraphics[width=0.32\linewidth]{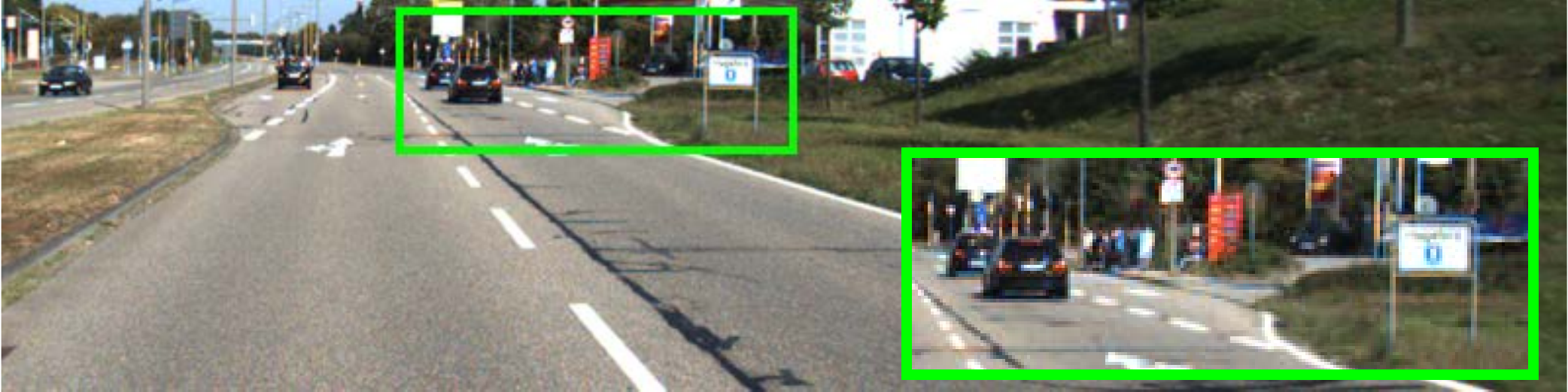} \\
\raisebox{2\height}{{\csmall (b)}}  &
\includegraphics[width=0.32\linewidth]{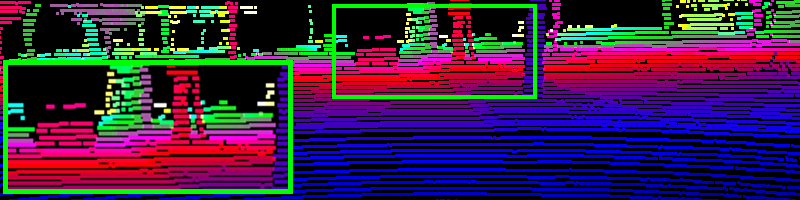} &
\includegraphics[width=0.32\linewidth]{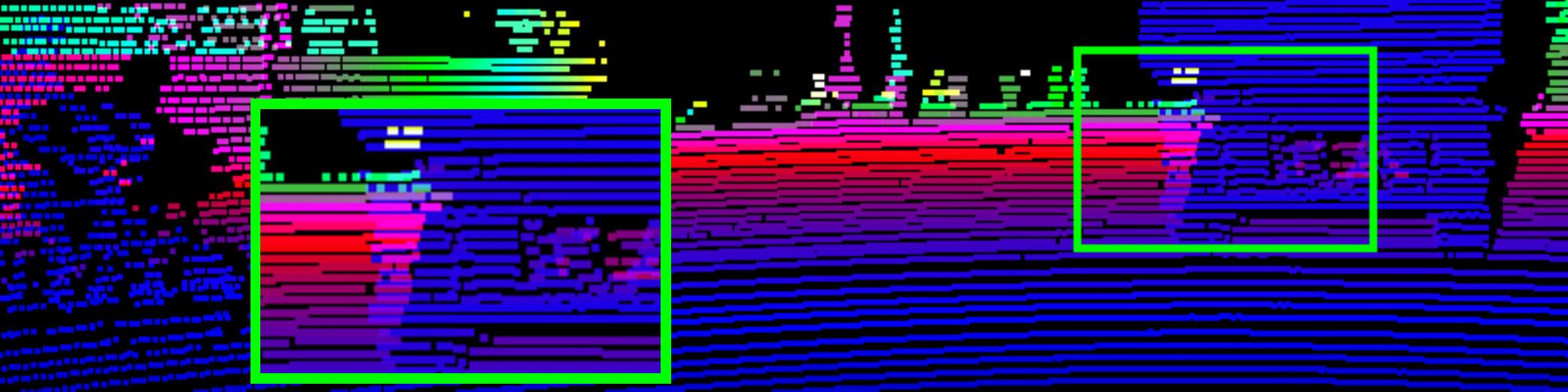} &
\includegraphics[width=0.32\linewidth]{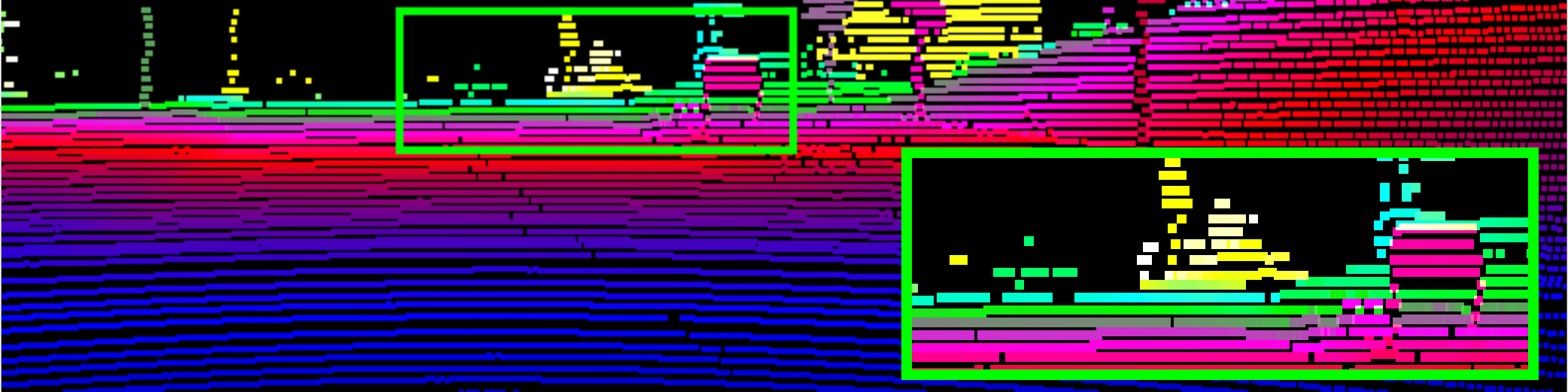} \\
\raisebox{2\height}{{\csmall (c)}}  &
\includegraphics[width=0.32\linewidth]{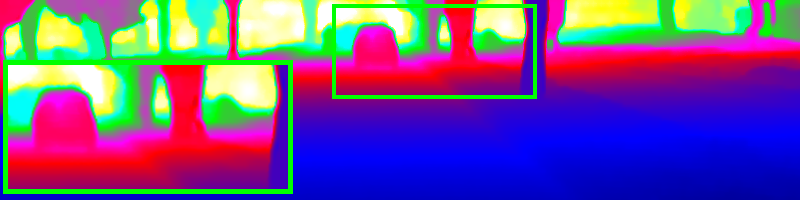} &
\includegraphics[width=0.32\linewidth]{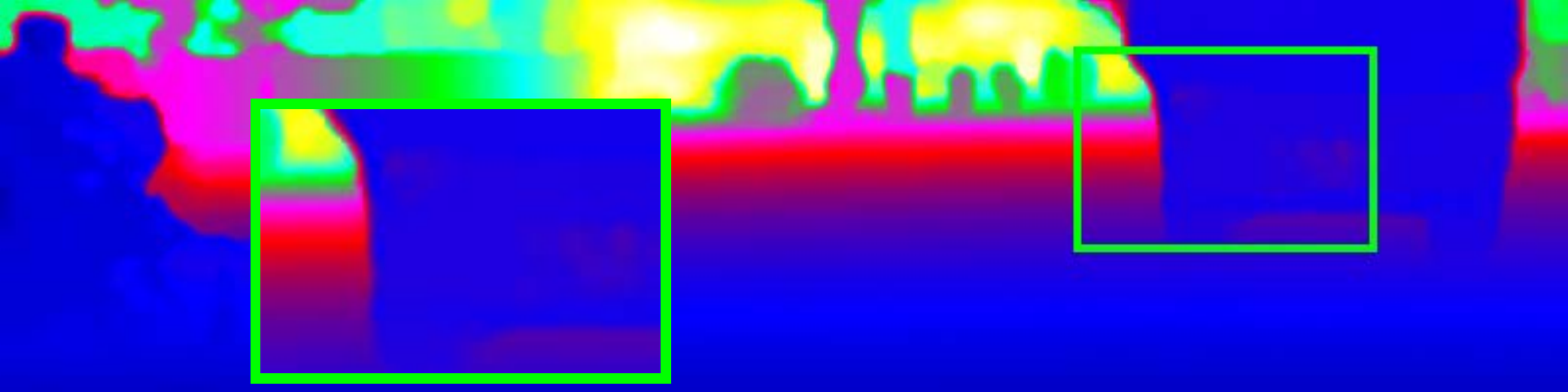} &
\includegraphics[width=0.32\linewidth]{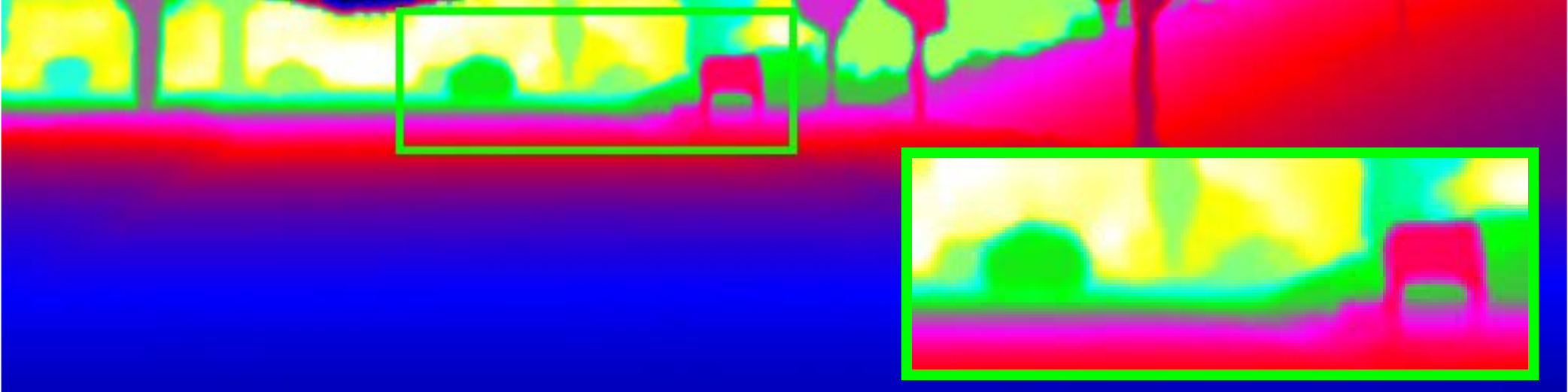} \\
\raisebox{2\height}{{\csmall (d)}}  &
\includegraphics[width=0.32\linewidth]{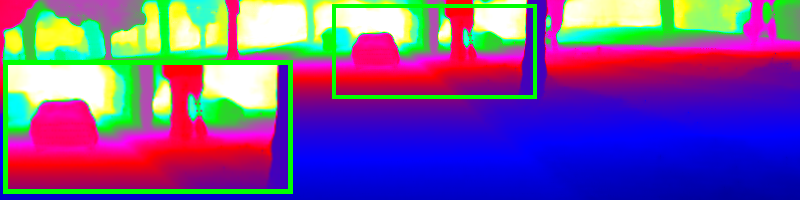} &
\includegraphics[width=0.32\linewidth]{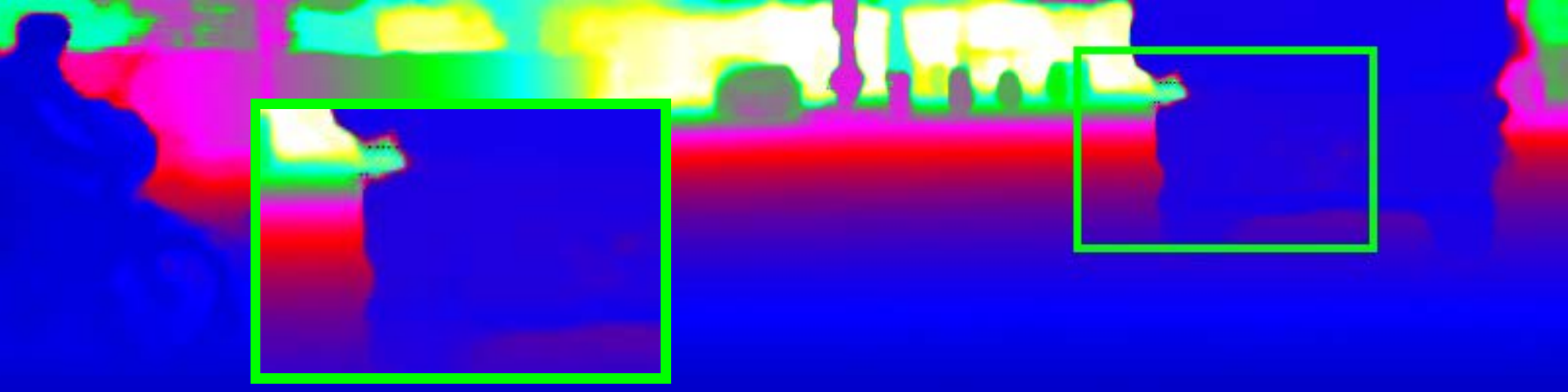} &
\includegraphics[width=0.32\linewidth]{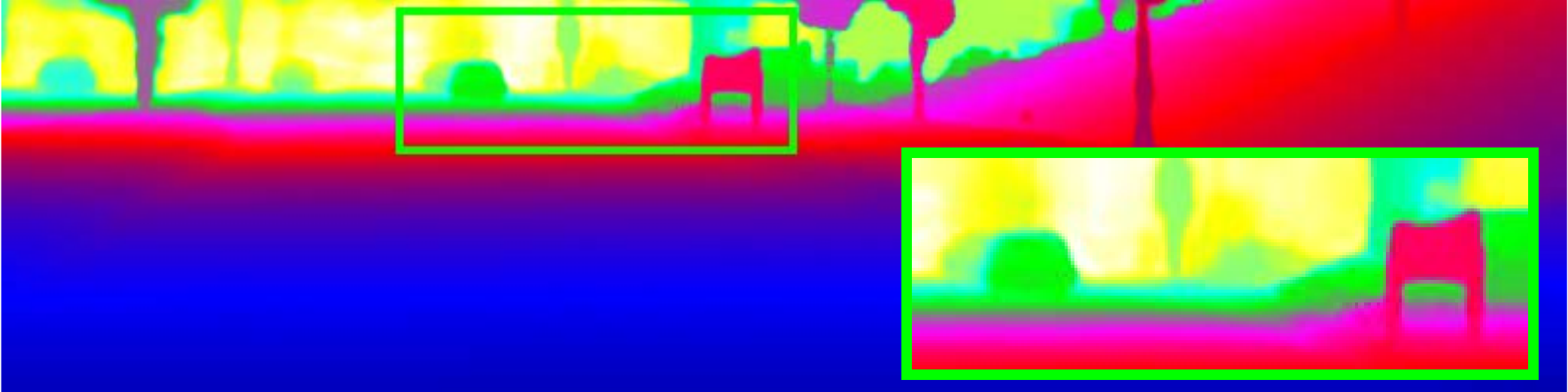} \\
\raisebox{2\height}{{\csmall (e)}}  &
\includegraphics[width=0.32\linewidth]{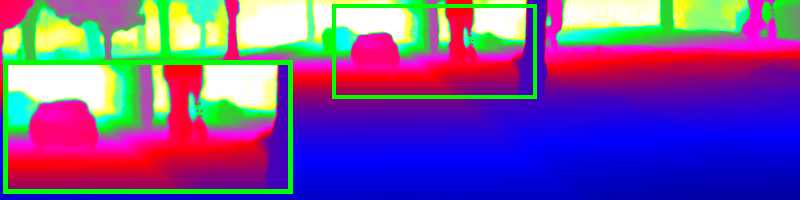} &
\includegraphics[width=0.32\linewidth]{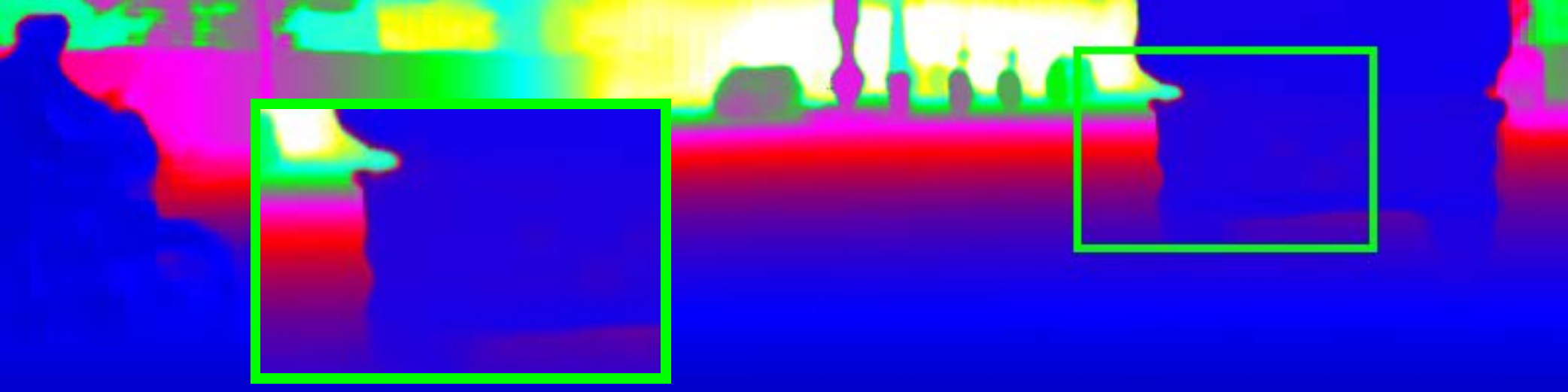} &
\includegraphics[width=0.32\linewidth]{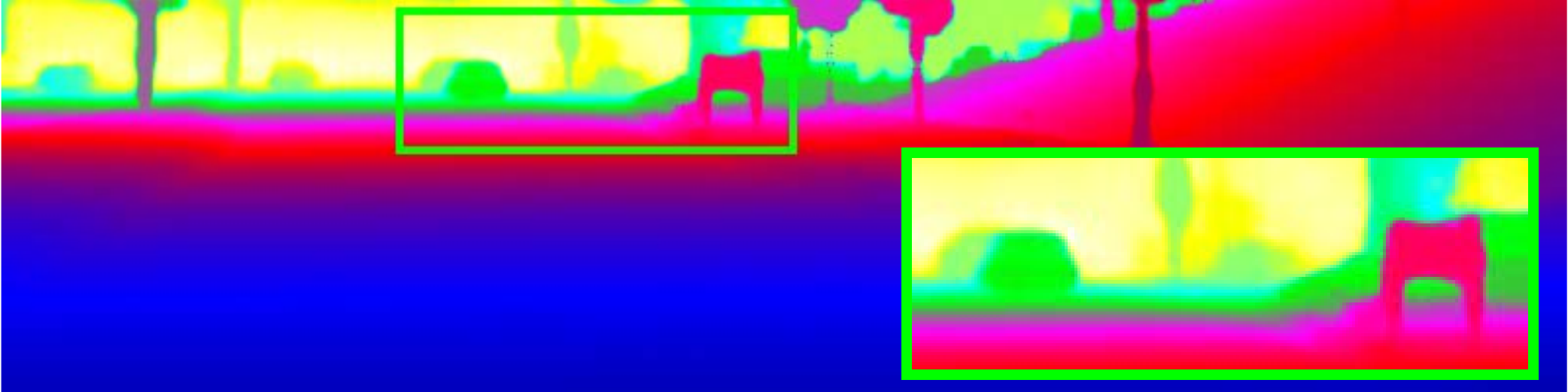} \\
\raisebox{2\height}{{\csmall (f)}}  &
\includegraphics[width=0.32\linewidth]{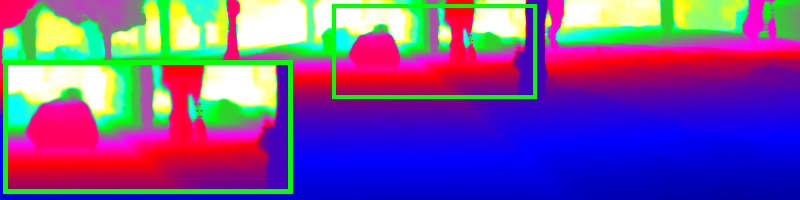} &
\includegraphics[width=0.32\linewidth]{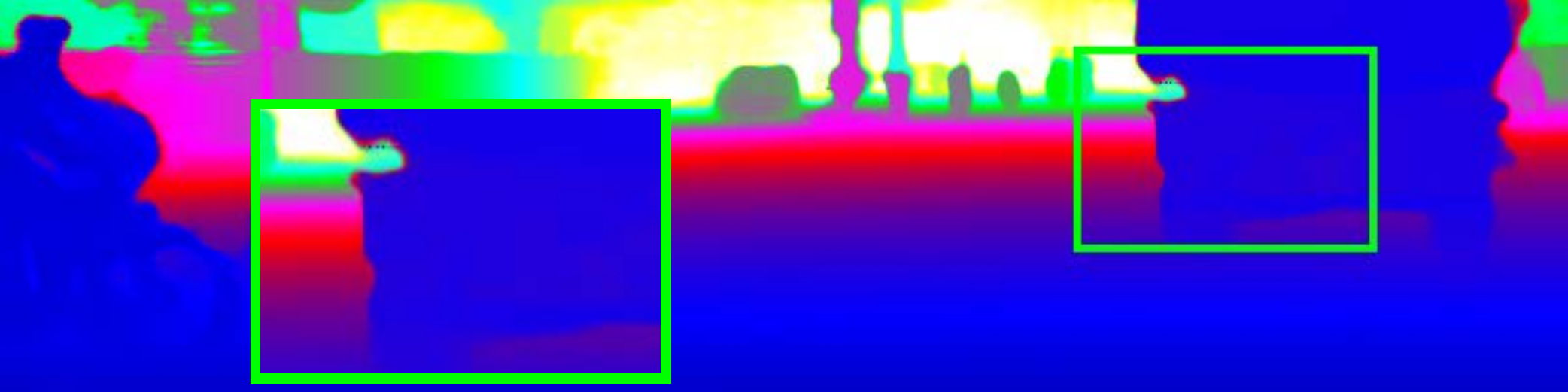} &
\includegraphics[width=0.32\linewidth]{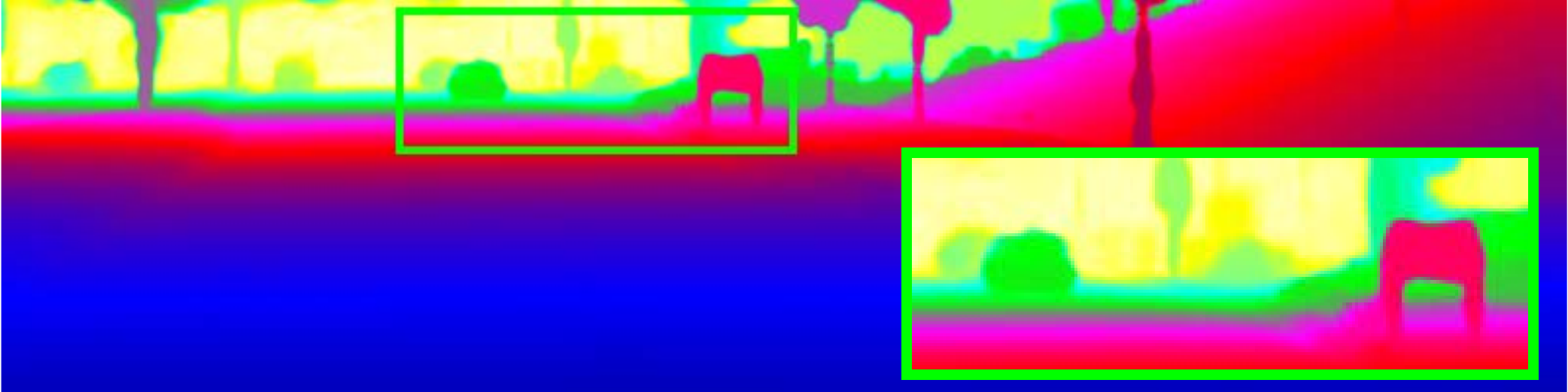} \\
\raisebox{2\height}{{\csmall (g)}}  &
\includegraphics[width=0.32\linewidth]{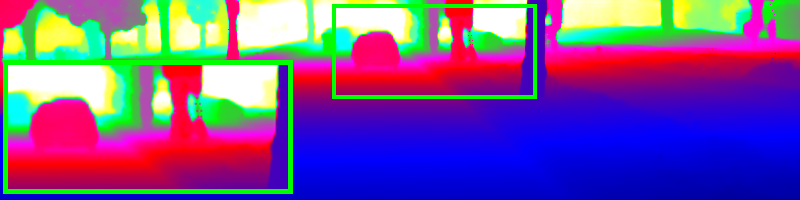} &
\includegraphics[width=0.32\linewidth]{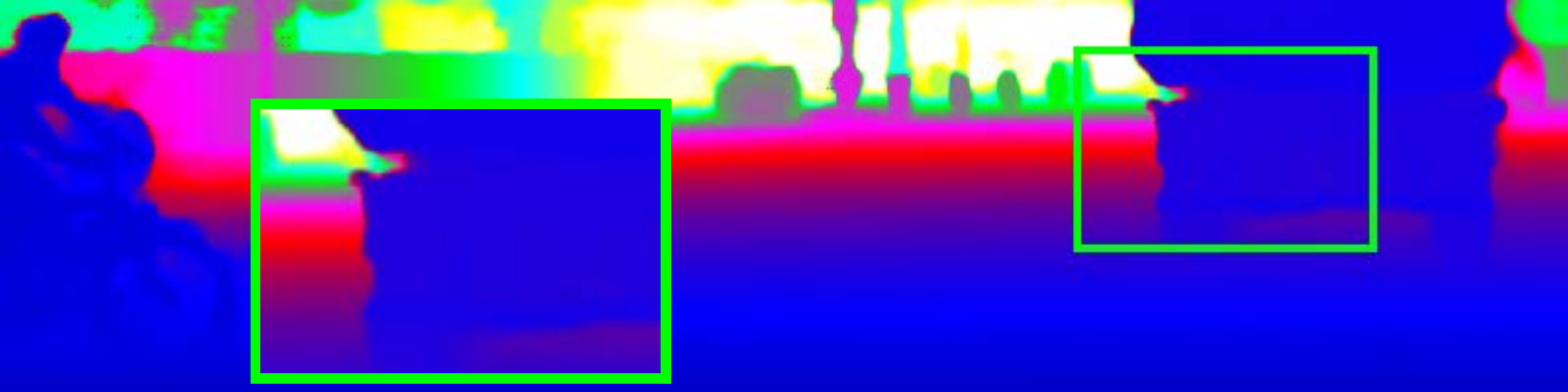} &
\includegraphics[width=0.32\linewidth]{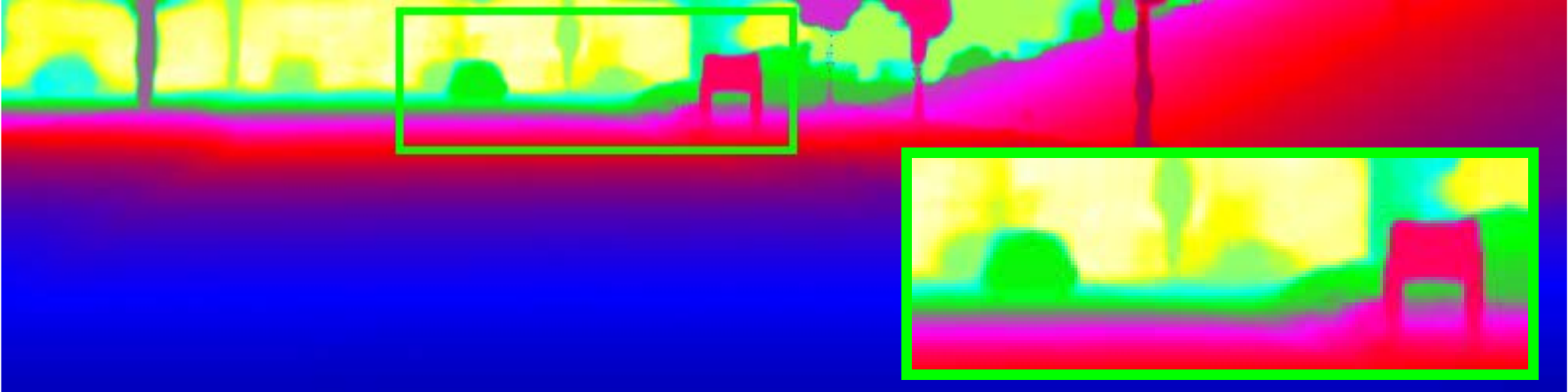} \\
\raisebox{2\height}{{\csmall (h)}}  &
\includegraphics[width=0.32\linewidth]{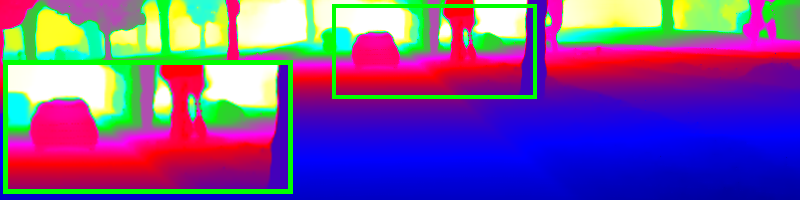} &
\includegraphics[width=0.32\linewidth]{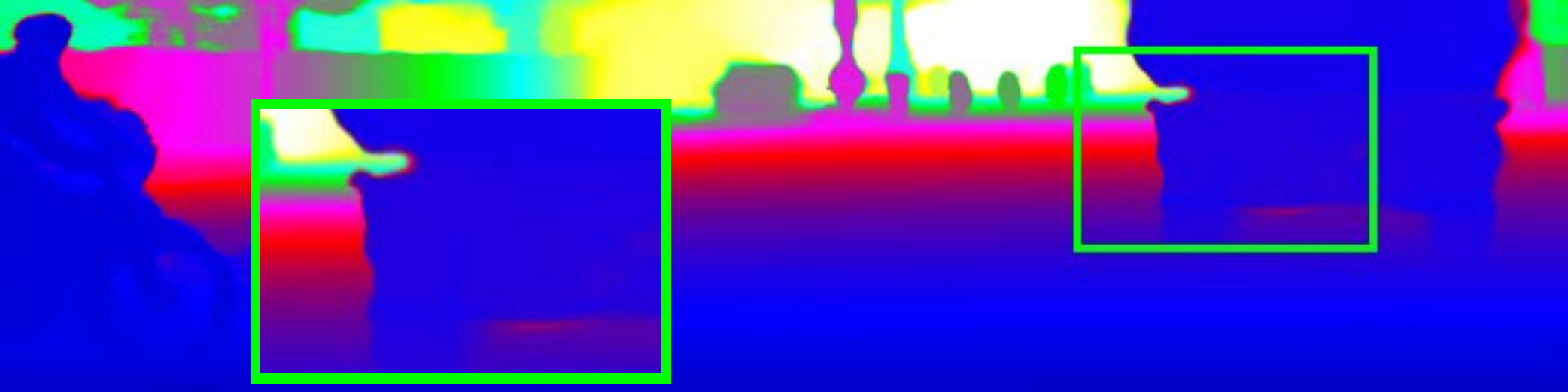} &
\includegraphics[width=0.32\linewidth]{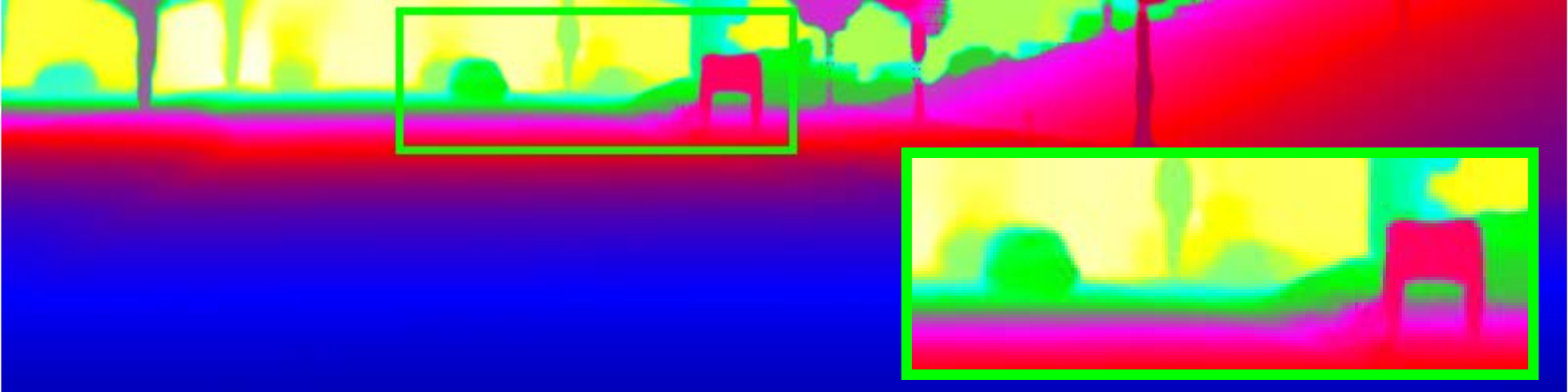} \\
\end{tabular}
\caption{\textbf{Depth completion results on the KITTI DC dataset~\cite{uhrig2017sparsity}}. (a) RGB, (b) Sparse depth, (c) CSPN~\cite{cheng2018depth}, (d) DepthNormal~\cite{xu2019depth}, (e) DeepLiDAR~\cite{qiu2019deeplidar}, (f) FuseNet~\cite{chen2019learning}, (g) CSPN++~\cite{cheng2019cspnpp}, (h) Ours. Note that sparse depth images are dilated for visualization.}
\label{fig:result_kittidc}
\end{center}
\end{figure*}

\Tabref{tab:quan_kitti} shows the quantitative evaluation of the KITTI DC dataset. 
Similar to the results obtained for the NYUv2, geometry-aware algorithms~\cite{xu2019depth,qiu2019deeplidar,chen2019learning,cheng2019cspnpp} perform better in general compared to geometry-agnostic methods~\cite{ma2018sparse,cheng2018depth}. 
Since LiDAR sensor noise (\textit{i.e.}, mixed foreground and background points as shown in \figref{fig:aff_conf_norm}) is inevitable, the predicted confidence is highly beneficial to eliminate the impact of the noise. 
DepthNormal~\cite{xu2019depth} utilizes confidence values as a mask for weighted summation during refinement. 
However, its confidence mask does not totally prevent incorrect values from propagating into neighboring pixels. 
%
On the contrary, the proposed confidence-incorporated affinity normalization effectively restricts the propagation of erroneous values during propagation. 
%
We note that the proposed method outperformed all the peer-reviewed methods in the KITTI online leaderboard when we submitted the paper.

\Figref{fig:result_kittidc} shows some examples of predicted dense depth with highlighted challenging areas. 
Those areas usually contain small structures near depth boundaries, which can be easily affected by the mixed-depth problem. 
Compared to the other methods (\figsref{fig:result_kittidc}(c)-(g)), our algorithm (\figref{fig:result_kittidc}(h)) handles those challenging areas better with the help of non-local neighbors.
%

\subsection{Ablation Studies}
\label{subsec:ablation}
We conducted ablation studies to verify the role of each component of our network, including non-local propagation, affinity normalization, and the confidence-incorporated propagation.
%
%
For all the experiments, we used a set of 10K images sampled from the KITTI DC training dataset for training and evaluated the performance on the full validation dataset. 
The network was trained for 20 epochs with center-cropped patches of $912{\times}228$ for fast training, and the batch size was set to 12. 
Other settings were set the same as those mentioned in~\secref{subsec:kitti_dc}.

\noindent \textbf{Non-Local Neighbors} \ 
\Figref{fig:abl_nl} visualizes some examples of non-local neighbors predicted by our algorithm. 
Compared to fixed-local neighbors, our predicted non-local neighbors have higher flexibility in the selection of neighbor pixels. 
In particular, non-local neighbors are selected from chromatically and geometrically relevant locations near the depth boundaries (\eg, same objects or planes). 
Moreover, we collected the statistics of the depth variance of neighboring pixels to show the relevance of the selected neighbors. 
On the KITTI DC validation set, the average depth variances for fixed-local and non-local neighbor configurations were $22.7$mm and $11.6$mm, respectively. 
The small variance of the non-local neighbor configuration demonstrates that the proposed method is able to select more relevant neighbors for propagation. 

%
%
%
The quantitative results obtained for the network with fixed-local $\mathcal{N}^{\mathrm{CS}}$ and that with non-local neighbors $\mathcal{N}^{\mathrm{NL}}$ are shown in \tabref{tab:abl_quan}. 
These networks were also tested with two normalization techniques: (1) with \absSum~(\tabref{tab:abl_quan}(b) and (g)), and (2) with \tgAbsSumStar~(\tabref{tab:abl_quan}(d) and (m)). 
The proposed method with non-local neighbors consistently outperformed that with fixed-local neighbors, demonstrating the superiority of the non-local framework.
%

\begin{table}[t]
\begin{tabular}{@{}c@{\hskip 0.02\linewidth}c}
\begin{minipage}[b]{0.36\linewidth}
    \begin{center}
        \includegraphics[width=0.98\linewidth]{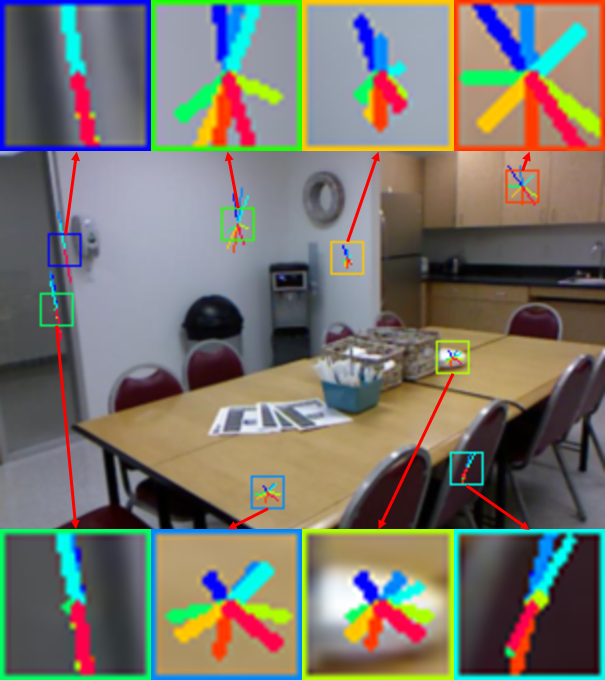}
        \captionof{figure}{\textbf{Examples of non-local neighbors predicted by our network}.
        }
        \label{fig:abl_nl}
    \end{center}
\end{minipage}
&
\begin{minipage}[b]{0.62\linewidth}
    \begin{center}
        {
        \scriptsize
        \renewcommand{\arraystretch}{1.15}
        \begin{tabular}{C{4mm}|C{12mm}|C{10mm}|C{20mm}|C{12mm}|C{10mm}}
        \hline
         & Neighbors & Affinity & Norm. & Conf. & RMSE (mm) \\ \hline \hline
        (a) & \multirow{4}{*}{$\mathcal{N}^{\mathrm{CS}}$} & \multirow{4}{*}{Learned} & \multirow{2}{*}{\absSum} & No & 908.4  \\ \cline{1-1}\cline{5-6}
        (b) &  &  &  & Yes & 891.6  \\ \cline{1-1}\cline{4-6}
        (c) &  &  & \multirow{3}{*}{\tiny \tgAbsSumStar} & No & 896.4  \\ \cline{1-1}\cline{5-6}
        (d) &  &  &   & \multirow{2}{*}{Yes} & 890.4  \\ \cline{1-3}\cline{6-6}
        (e) & \multirow{9}{*}{$\mathcal{N}^{\mathrm{NL}}$} & Color &  &  & 930.3  \\ \cline{1-1}\cline{3-6}
        (f) &  & \multirow{8}{*}{Learned} & \multirow{2}{*}{\absSum} & No & 903.1  \\ \cline{1-1}\cline{5-6}
        (g) &  &  &  & \multirow{3}{*}{Yes} & 889.5  \\ \cline{1-1}\cline{4-4}\cline{6-6}
        (h) &  &  & \absSumStar &  & 886.0  \\ \cline{1-1}\cline{4-4}\cline{6-6}
        (i) &  &  & \tanhC &  & 886.4  \\ \cline{1-1}\cline{4-6}
        (j) &  &  & \multirow{4}{*}{\tiny\tgAbsSumStar} & No & 891.3  \\ \cline{1-1}\cline{5-6}
        (k) &  &  &  & Binary & 892.9  \\ \cline{1-1}\cline{5-6}
        (l) &  &  &  & Weighted & 884.8  \\ \cline{1-1}\cline{5-6}
        (m) &  &  &  & Yes & \textbf{884.1}  \\ \hline
        \end{tabular}
        }
        \captionof{table}{\textbf{Quantitative evaluation on the KITTI DC validation set~\cite{uhrig2017sparsity} with various configurations}. Please refer to the text for details.}
        \label{tab:abl_quan}
    \end{center}
\end{minipage}
\end{tabular}
\end{table}

\noindent \textbf{Affinity Normalization and Confidence Incorporation} \ 
To validate the proposed affinity normalization algorithm, we compare it with three different affinity normalization methods (\textit{cf.},~\secref{sec:aff}). 
\Tabref{tab:abl_quan}(g)-(i), and (m) assessed the performance using the same network but different affinity normalization methods. 
The model with \absSum~does not perform well due to the limited range of affinity combinations, as shown in \figref{fig:aff_norm}(a). 
%
When relaxing the normalization condition while maintaining the stability condition (\absSumStar), the performance was improved thanks to the wider area of feasible affinity space and better affinity distribution (\figref{fig:aff_norm}(b)). 
%
\tanhC~strengthens the stability condition without explicit normalization. 
However, as shown in \figref{fig:aff_norm}(c), the resulting affinity values reside in a smaller affinity space (\ie, in a $K$-dimensional hypercube with edge size $2/K$); therefore, it achieved a slightly worse performance compared to \absSumStar. 
The proposed \tgAbsSumStar~was able to alleviate this limitation with a learnable normalization parameter $\gamma$. 
The learned $\gamma$ compromises between \absSumStar~and \tanhC, and can boost the performance. 
%
Note that the final $\gamma$ values (initialized with $\gamma = K = 8$) trained on the NYUv2 (\secref{subsec:exp_nyu}) and the KITTI DC (\secref{subsec:kitti_dc}) datasets were 5.2 and 6.3, respectively. 
%
This observation indicates that the optimal $\gamma$ varies based on the training environment. 
%


We also compared the performance of the network with and without confidence, to verify the importance of confidence incorporation. 
%
%
In addition, we tested two alternative confidence-aware networks (1) by generating a binary mask from confidence with a threshold of $0.5$ and (2) with the weighted summation approach of DepthNormal~\cite{xu2019depth}, and applying each method during the propagation to eliminate the effect of outliers. 
%
The comparison results are shown in \tabref{tab:abl_quan}(j)-(m). 
%
The proposed confidence-incorporated affinity normalization (\tabref{tab:abl_quan}(m)) outperforms the others due to its capability of suppressing propagation from unreliable pixels.
%
The mask-based (\tabref{tab:abl_quan}(k)) and weighted summation (\tabref{tab:abl_quan}(l)) approaches show worse performance compared to that of ours, indicating that the hard-thresholding and weighted summation approaches are not optimal for encouraging propagation from relevant pixels but suppressing that from irrelevant pixels. 
%
Note that the proposed confidence-incorporated approach is effective for both the network with $\mathcal{N}^{\mathrm{NL}}$ and that with $\mathcal{N}^{\mathrm{CS}}$ (\tabref{tab:abl_quan}(a)-(d)). 
These results demonstrate the effectiveness of our confidence incorporation.

\begin{table*}[t]
\begin{center}
{\ctiny
\renewcommand{\arraystretch}{1.2}
\begin{tabular}{C{18mm}C{11mm}C{11mm}C{12mm}C{11mm}C{19mm}C{19mm}C{11mm}}
 \hline
 Method & CSPN~\cite{cheng2018depth} & DDP~\cite{yang2019dense} & NConv~\cite{eldesokey2019confidence} & S2D~\cite{ma2018sparse} & DepthNormal~\cite{xu2019depth} & DeepLiDAR~\cite{qiu2019deeplidar} & Ours \\ 
 \hline
 \# Params. (M) & 17.41 & 28.99 & 0.36 & 42.82 & 28.99 & 53.44 & 25.84 \\ \hline
\end{tabular}
}
\caption{
\textbf{Comparison of the number of network parameters}.
Note that only methods with publicly available implementations~\cite{cheng2018depth,yang2019dense,eldesokey2019confidence,ma2018sparse,xu2019depth,qiu2019deeplidar} are included.
}
\label{tab:num_param}
\end{center}
\end{table*}

\noindent \textbf{Further Analysis} \ 
To verify the importance of learned affinities, we further evaluated the proposed method with conventional affinities calculated based on the Euclidean distance between color intensities.
As shown in \tabref{tab:abl_quan}(e) and (m), the network using learned affinities performed much better than the network using the hand-crafted one. 
%
In addition, we provide the number of network parameters of the compared methods in \tabref{tab:num_param}. 
%
%
The proposed method achieved superior performance with a relatively small number of network parameters. 
%
Please refer to the supplementary material for additional experimental results, visualizations, and ablation studies.

\section{Conclusion}
We have proposed an end-to-end trainable non-local spatial propagation network for depth completion. 
The proposed method gives high flexibility in selecting neighbors for propagation, which is beneficial for accurate propagation, and it eases the affinity learning problem. 
Unlike previous algorithms (\ie, fixed-local propagation), the proposed non-local spatial propagation efficiently excludes irrelevant neighbors and enforces the propagation to focus on a synergy between relevant ones. 
%
In addition, the proposed confidence-incorporated learnable affinity normalization encourages more affinity combinations and minimizes harmful effects from incorrect depth values during propagation. 
%
Our experimental results demonstrated the superiority of the proposed method. \newline

\noindent\textbf{Acknowledgement}\ This work was partially supported by the National Information Society Agency for construction of training data for artificial intelligence (2100-2131-305-107-19).

%
%
\bibliographystyle{splncs04}
\bibliography{egbib}
\end{document}